\PassOptionsToPackage{table}{xcolor}
\documentclass[fleqn,11pt]{wlscirep}

\usepackage[T1]{fontenc}
\usepackage{setspace}
\usepackage{multirow}
\usepackage{array}

\newcolumntype{R}[1]{>{\raggedleft\arraybackslash}p{#1}}
\newcolumntype{L}[1]{>{\raggedright\arraybackslash}p{#1}}
\newcolumntype{P}[1]{>{\centering\arraybackslash}p{#1}}

\title{Deep learning-based detection of cessation of breathing in pre-term infants}
\author[1]{Dineo Serame}
\author[1]{Lionel Tarassenko}
\author[1]{Mauricio Villarroel}
\affil[1]{Institute of Biomedical Engineering, Department of Engineering Science,  University of Oxford, UK}
\affil[*]{mauricio.villarroel@eng.ox.ac.uk}

\begin{abstract}
Apnoea of prematurity is characterised by recurrent episodes of cessation of breathing and remains difficult to detect reliably using routinely monitored physiological signals in the Neonatal Intensive Care Unit (NICU). Bedside monitors generate alarms based on respiratory rate and oxygen saturation thresholds. However, such rule-based approaches can exhibit high false positive rates and fail to capture short or irregular events, contributing to frequent non-actionable alarms. Improving automated detection using existing clinical signals could enhance discrimination of clinically meaningful events without increasing infant burden or requiring additional sensing hardware.

\paragraph{}
We evaluated deep learning-based detection of apnoea-related Cessation Of BrEathing (COBE) events using routinely acquired NICU signals. We analysed impedance pneumography (IP), electrocardiography (ECG), and photoplethysmography (PPG) signals for approximately 430 hours of NICU recordings from 24 pre-term infants. Three independent reviewers annotated cessation of breathing events, yielding a dataset of 346 COBE and 608 non-COBE events. Using this dataset, we compared a shallow convolutional neural network (CNN), residual networks (ResNets), and a ConvNeXt architecture. Performance was assessed on an independent held-out test set.

\paragraph{}
Across all architectures, detection performance was more strongly influenced by signal modality than by architectural complexity. Unimodal IP-based models (86.8\% -- 88.0\% balanced accuracy on the independent test set) consistently outperformed ECG-derived (62.6\% -- 69.7\%) and PPG-derived (65.1\% -- 66.4\%) respiratory surrogates. Multimodal fusion yielded modest improvements over IP alone. The best-performing configuration, a ConvNeXt model combining IP and PPG inputs, achieved a balanced accuracy of 88.7\% on the independent test set with an F1 score of 0.75.

\paragraph{}
These results show that deep learning models applied to routinely monitored NICU signals achieve reliable detection of COBE events. Signal modality has a greater influence on performance than architectural complexity in data-constrained neonatal monitoring settings.
\end{abstract}

\begin{document}
\setlength{\belowdisplayskip}{10pt} \setlength{\belowdisplayshortskip}{5pt}
\setlength{\abovedisplayskip}{1pt} \setlength{\abovedisplayshortskip}{3pt}
\raggedbottom
\maketitle
\thispagestyle{empty}
\newpage

\section{Introduction}
\label{intro}
Apnoea of prematurity (AOP) is a common and clinically significant complication in pre-term infants~\cite{seppa2022sleep}. It is characterised by recurrent episodes of Cessation Of BrEathing (COBE) exceeding 20 seconds, or shorter pauses ($\le$ 10 seconds) accompanied by bradycardia (heart rate <100 beats per minute) or oxygen desaturation ($SpO_{2} < 80\%$)~\cite{zhao2011apnea,bester2018study}. AOP arises primarily from immature neurological and respiratory control mechanisms~\cite{kondamudi2023infant}, rendering pre-term infants vulnerable to sudden failures of central respiratory drive or unstable airway mechanics. Episodes may be central, obstructive, or mixed in origin~\cite{almutairi2021classification,martin2004apnoea}, and are frequently accompanied by low levels of oxygen in the blood. Recurrent or prolonged events are associated with adverse outcomes, including impaired cerebral oxygenation, neurodevelopmental impairment, and increased risk of mortality~\cite{mohr2015very,williamson2021apnoea}.

\paragraph{}
Adult sleep apnoea is typically quantified using the Apnoea-Hypopnea Index (AHI), defined as the number of events per hour of sleep~\cite{xavier2019temporal,korompili2021detecting}. In contrast, AOP presents as short-duration, clinically dynamic episodes requiring timely detection. This distinction underscores the need for new monitoring strategies tailored to the acute and physiologically unstable respiratory events characteristic of pre-term infants.

\paragraph{}
Polysomnography (PSG) remains the reference standard for diagnosing apnoea, recording airflow, respiratory effort, cardiac activity, and oxygen saturation to comprehensively characterise events~\cite{bertoni2019towards,zou2023new}. However, PSG is expensive, requires specialised staff, and is impractical for continuous bedside monitoring in the Neonatal Intensive Care Unit (NICU)~\cite{de2004automated}. Consequently, detection of COBE in pre-term infants relies on routinely monitored physiological signals. In clinical practice, cardiorespiratory monitoring and pulse oximetry are used to assess respiratory and cardiovascular status~\cite{pullano2017medical}. Electrocardiography (ECG) is used to record cardiac electrical activity, from which bedside monitors derive heart rate and detect bradycardia~\cite{almutairi2021detection}. The same ECG electrodes can be used for recording the impedance pneumography (IP) signal, which measures changes in thoracic impedance associated with lung aeration to estimate respiratory rate~\cite{bawua2021review}. Pulse oximetry provides continuous estimation of peripheral oxygen saturation ($SpO_{2}$) using photoplethysmographic (PPG) measurement of pulsatile arterial blood~\cite{poets2003pulse}. These signals are integrated into bedside monitors that generate alarms when respiratory rate, heart rate, or oxygen saturation cross predefined thresholds. 

\paragraph{}
Beyond direct respiratory monitoring through IP, respiratory information can also be inferred indirectly from cardiac and perfusion signals. In ECG, respiration modulates heart rate variability and beat morphology, enabling estimation of ECG-derived respiration (EDR)~\cite{zarei2020automatic,8681397,charlton2017breathing,ponsiglione2025comparison}. Similarly, respiratory modulation in PPG arises from changes in venous return and peripheral blood volume.  However, these surrogate respiratory signals are sensitive to motion artefacts and low signal-to-noise conditions. In pre-term infants, immature autonomic regulation and irregular breathing patterns may further reduce the stability of respiration-related variations in ECG and PPG signals~\cite{joshi2019cardiorespiratory}. In addition, PPG-derived respiration may become unreliable during periods of low peripheral perfusion~\cite{charlton2017breathing}. Combining complementary information from IP, ECG, and PPG may therefore improve robustness of COBE detection.

\paragraph{}
Despite their complementary roles, these monitoring modalities have important limitations. IP is susceptible to motion artefacts, electrode displacement, and cardiac interference, which can obscure respiratory pauses or generate false alarms~\cite{bawua2021review,bertoni2019towards}. Pulse oximeters commonly apply temporal signal averaging to reduce noise and stabilise displayed oxygen saturation values. However, this signal averaging can reduce sensitivity to rapid physiological changes and delay detection of transient desaturation events~\cite{poets2003pulse,vagedes2024averaging}. Additional delays may arise because decreases in arterial oxygen saturation take several seconds to be detected at peripheral measurement sites~\cite{poets2003pulse}. Consequently, brief desaturation episodes associated with COBE may not immediately trigger bedside alarms, while motion artefacts or poor sensor contact can generate non-actionable alerts. High false alarm rates in NICUs therefore contribute to alarm fatigue and may reduce responsiveness to genuine physiological events~\cite{ostojic2020reducing,hravnak2018call}.

\paragraph{}
Automated apnoea detection has been extensively studied in adult populations, where machine learning approaches have been applied to ECG, airflow, and oximetry-derived signals. Conventional methods have employed support vector machines, decision trees, and random forests on handcrafted features~\cite{bsoul2010apnea,xie2012real,song2015obstructive,garcia2022artifacts}. Recent work has adopted deep learning architectures, including convolutional neural networks (CNNs), residual networks (ResNets), and long short-term memory (LSTM) models, often incorporating multimodal data fusion to improve robustness~\cite{almutairi2021detection,urtnasan2018automated,fayyaz2024multimodal,tao2025multimodal}. Many of these studies are evaluated on standardised datasets such as the PhysioNet Apnea-ECG database, in which annotations are mapped to fixed one-minute epochs and classified as either normal or disordered breathing~\cite{penzel2000apnea}. This minute-level segmentation is suited to quantifying overall breathing disturbance but does not extend directly to neonatal populations, where respiratory events are shorter, physiologically unstable, and occur within signals shaped by developmental immaturity.

\paragraph{}
In contrast, neonatal studies have primarily used classical machine learning approaches, including quadratic classifiers, Gaussian mixture models, support vector machines, and autoencoders applied to handcrafted physiological features~\cite{williamson2011using,williamson2013individualized,shirwaikar2019optimizing,adjei2021new,varisco2024detecting}. More recently, deep learning approaches have begun to emerge in neonatal respiratory event prediction and classification, including neural additive models, CNNs, and recurrent neural network-based architectures~\cite{vetter2024neonatal,krupa2025automated}. However, progress in this area is constrained by the limited availability of large annotated datasets from pre-term infants. For example, Krupa \textit{et al.}~\cite{krupa2025automated} developed CNN- and CNN-BiLSTM-based models to classify apnoea and hypoxia events from synthetically generated neonatal respiratory signals, highlighting the challenges posed by the scarcity of real-world neonatal datasets. Although many of the studies adopt similar clinical definitions of apnoea, their implementation for machine learning, particularly window length, overlap, and labelling strategies, varies considerably, complicating direct comparison of reported performance. For example, some neonatal apnoea prediction models extract features from 60-second sliding windows to identify events several minutes in advance~\cite{williamson2011using}, whereas others use shorter or non-overlapping windows for near-real-time apnoea detection. In addition, existing neonatal work has typically evaluated only a narrow range of model architectures or signal configurations, without systematically examining how signal modality and architectural complexity influence detection performance. 

\paragraph{}
To address these limitations, we investigate the application of deep learning architectures to COBE detection in pre-term infants using routinely monitored physiological signals acquired in the NICU. We evaluate a shallow CNN, ResNet architectures, and a ConvNeXt model across unimodal and multimodal combinations of IP, ECG, and PPG signals. This evaluation examines how architectural complexity and signal modality influence COBE detection performance in pre-term infants.

\section{Methods}
\label{methods}
\subsection{Clinical study}
We conducted a clinical study as part of a collaboration between the Oxford University Hospitals NHS Foundation Trust and the Oxford Biomedical Research Centre. The study complied with institutional and governmental regulations and was approved by the South Central - Oxford Research Ethics Committee under reference number 13/SC/0597.

\paragraph{}
Pre-term infants (<37 weeks gestation) were nursed in the high-dependency area of the NICU at the John Radcliffe Hospital in Oxford. Infants were monitored for up to 7 days during daytime hours, without interfering with regular patient care. Further details of the study protocol are described in Villarroel \textit{et al.}~\cite{villarroel2019non}.

\paragraph{}
Infants were recruited according to the British Association of Perinatal Medicine’s Categories of Care (2011). Inclusion criteria were gestational age <37 weeks, requirement for high-dependency care, and continuous monitoring of heart rate, respiratory rate, and oxygen saturation. Infants with life-threatening conditions requiring intensive care were excluded. Parents received written and verbal study information and provided written informed consent. All data were anonymised and securely stored, with infants identified only by a study-specific participant number.

\paragraph{}
Physiological signals were recorded continuously using standard patient monitoring equipment. The study included 30 pre-term infants (mean gestational age 31.1 $\pm$ 1.8 weeks) monitored across 90 recording sessions, yielding approximately 429.5 hours of reference physiological recordings. Heart rate and respiratory rate were measured using a Philips IntelliVue MX800 patient monitor (Philips, Amsterdam, Netherlands), and $SpO_{2}$ was recorded using a Masimo Vuelink IntelliVue pulse oximeter module (Masimo, California, USA). Table~\ref{fs} summarises the recorded physiological signals.

\begin{table}[h]
\centering
\caption{Physiological parameters recorded by the Philips monitor and the Masimo pulse oximeter}
\label{fs}
\scalebox{1.0}{
\begin{tabular}{>{\raggedright\arraybackslash}p{3.0cm}
                >{\raggedright\arraybackslash}p{10.2cm}
                R{3.0cm}}
\multicolumn{1}{c}{\textbf{Name (units)}} &
\multicolumn{1}{c}{\textbf{Description}} &
\multicolumn{1}{c}{\textbf{Sampling rate (Hz)}} \\
\hline
ECG (mV) & 1-lead electrocardiography signal & 500 \\
PPG (a.d.u) & Photoplethysmography signal & 125 \\
IP (ohms) & Impedance pneumography signal & 62.5 \\
RR (breaths/min) & Respiratory rate computed by the Philips monitor (displayed value) & 1 \\
HR (beats/min) & Heart rate computed by the Philips monitor & 1 \\
$SpO_{2}$ (\%) & Oxygen saturation computed by the Masimo pulse oximeter & 1 \\
\bottomrule
\end{tabular}}
\end{table}

\subsection{Creating a labelled dataset for COBE detection}
COBE events were identified using established clinical criteria for apnoea, defined as either a pause in breathing lasting (i) at least 20 seconds, or (ii) at least 10 seconds accompanied by bradycardia ($HR$ < 100 beats/min) or oxygen desaturation ($SpO_{2}$ < 80\%)~\cite{zhao2011apnea,bester2018study}.

\paragraph{}
To reduce the volume of data requiring manual review, an automated screening stage was implemented in MATLAB to identify candidate events for annotation. The algorithm identified periods where $SpO_{2}$  < 80\% for at least 10 seconds, corresponding to the oxygen desaturation threshold used in the clinical definition of apnoea~\cite{bester2018study}. Consecutive desaturation episodes occurring within 20 seconds of each other were merged into a single candidate event to avoid splitting prolonged desaturation episodes into multiple events because of brief transient recoveries in $SpO_{2}$. This approach was motivated by observations that desaturation and apnoea-related events may occur in temporal clusters during periods of physiological instability rather than as entirely independent events~\cite{joshi2016pattern}. Merging was performed only when at least 75\% of the combined interval remained below the desaturation threshold, such that the merged segment predominantly represented sustained desaturation rather than isolated transient decreases in oxygen saturation.
 
\paragraph{}
The screening procedure identified 620 candidate desaturation-associated segments. To ensure balanced representation of both desaturation-associated and non-desaturation respiratory patterns, an additional 620 non-desaturation segments were randomly sampled from the same recording sessions for review.  

\paragraph{}
Three reviewers (one clinician and two biomedical engineers) independently labelled the dataset using a custom MATLAB annotation interface. They inspected 5-minute windows centred on each candidate event, with access to $RR$ and $SpO_{2}$ as primary signals, and ECG, PPG, and IP waveforms as supplementary references. An example of the annotation interface is provided in Appendix~\ref{annot_gui}. 

\paragraph{}
Segments containing periods of missing physiological data due to sensor disconnection or for which the three reviewers could not reach consensus were excluded from further analysis. The remaining segments were categorised into four classes according to respiratory rate and desaturation criteria: (i) COBE with desaturation, (ii) normal breathing with desaturation, (iii) COBE without desaturation, and (iv) normal breathing without desaturation. COBE with desaturation was defined as $RR \le 20$ breaths/min for at least 10 seconds occurring before or during a desaturation episode ($SpO_{2}$ < 80\%), while COBE without desaturation was defined as $RR \le 20$ breaths/min for at least 20 seconds without an associated desaturation event.  Inter-reviewer agreement was substantial for desaturation-associated events (Fleiss' $\kappa = 0.80$) and moderate for non-desaturation respiratory pauses ($\kappa = 0.57$). The decision workflow used to standardise annotation is shown in Appendix~\ref{annotate_process}. 

\paragraph{}
Following annotation, shorter fixed-length segments were extracted around each identified event for machine learning analysis. Each segment was defined as an 80-second window comprising 60 seconds preceding the onset of the desaturation event and 20 seconds following it, providing pre-event physiological context and capturing respiratory activity surrounding the desaturation. These segments were then partitioned into overlapping 20-second windows with a 10-second step size, yielding seven windows per segment. A 20-second window length was selected to align with established clinical definitions of apnoea ($\geq$ 20 seconds) while ensuring that shorter clinically relevant pauses ($\geq$ 10 seconds with desaturation) could be fully contained within at least one analysis window. A 10-second step size introduced 50\% overlap between adjacent windows, reducing boundary effects and ensuring continuous temporal coverage. Windows containing at least 10 seconds of $RR$ below 20 breaths/min were assigned a positive label. All remaining windows were labelled as negative. 

\paragraph{}
After exclusion of segments with sensor disconnection or lack of reviewer consensus, the final dataset comprised 346 COBE events and 608 non-COBE segments derived from 24 of the 30 enrolled infants. The 6 excluded infants did not contribute annotated events meeting the inclusion criteria. A summary of the final annotated dataset is shown in table~\ref{dataset}.

\begin{table}[h]
\centering
\caption{Composition of the final annotated dataset, including the number of annotated segments in each class, the total number of analysis windows, and the number of contributing infants.}
\label{dataset}
\scalebox{1.0}{
\begin{tabular}{>{\raggedright\arraybackslash}p{6.5cm}
                R{3.5cm}}
\multicolumn{1}{c}{\textbf{Dataset component}} &
\multicolumn{1}{c}{\textbf{Count}} \\
\hline
COBE with desaturation & 283 \\
COBE without desaturation & 63 \\
Normal with desaturation & 193 \\
Normal without desaturation & 415 \\
\hline
Total analysis windows & 6,678 \\
Contributing infants & 24 \\

\hline
\end{tabular}}
\end{table}

\subsection{Respiratory signal extraction}
For machine learning analysis, three respiratory input signals were obtained from the recorded physiological waveforms. The filtered IP waveform was used directly as a measure of respiratory effort. An ECG-derived respiration (EDR) signal was generated from respiratory modulation of ECG R-peak amplitudes, while respiratory-induced variations in the PPG signal were represented using a continuous envelope derived from successive PPG peaks. These signals provided complementary representations of respiratory physiology from IP, ECG, and PPG recordings. Full details of signal extraction and preprocessing are provided in Appendix~\ref{appendix:supp_signals}.

\subsection{Deep learning architectures for COBE detection}
To evaluate the influence of network architecture on COBE detection performance, three convolutional model families were implemented: a shallow 1D CNN similar to Urtnasan \textit{et al.}~\cite{urtnasan2018automated},  ResNets~\cite{carter2023deep}, and ConvNeXt~\cite{Liu_2022_CVPR} architectures. Each architecture was evaluated using unimodal inputs (IP, EDR, and PPG envelope) and multimodal configurations. For early fusion, signals were combined at the input level before feature extraction. For late fusion, each signal was processed through a separate network branch before the learned feature representations were concatenated for classification. All multimodal models were trained end-to-end using a shared classification loss.

\subsubsection{Convolutional Neural Network (CNN) baseline} 
\begin{figure}[!ht]
\centering
\includegraphics[width = 17.5 cm]{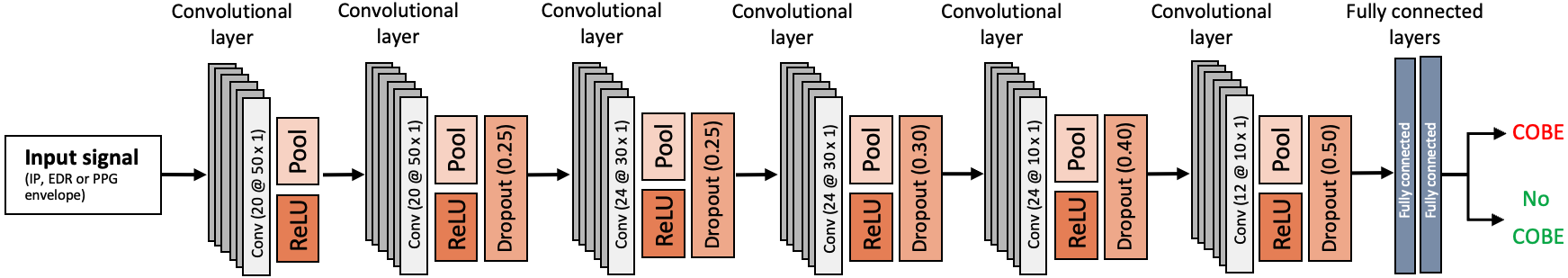}
\caption {Architecture of the CNN model modified for automated COBE episode detection. The model processes 1D physiological signals (IP, EDR, or PPG envelope) through a series of convolutional layers with ReLU activations, average pooling, and dropout, followed by fully connected layers for binary classification. The same architecture is applied to individual signals and extended to multimodal configurations using fusion strategies.}
           \label{urt_cnn}
\end{figure}

A one-dimensional CNN was implemented as a baseline architecture for comparison. The model was based on the six-layer CNN proposed by Urtnasan \textit{et al.}~\cite{urtnasan2018automated}, originally developed for adult apnoea detection using ECG recordings. The network comprises six convolutional layers with ReLU activations and pooling operations, followed by two fully connected classification layers. To adapt the architecture to neonatal physiological signals (IP, EDR, and PPG envelope), kernel sizes were modified across successive layers (50$\times$1, 50$\times$1, 30$\times$1, 30$\times$1, 10$\times$1, and 10$\times$1). Additional modifications included the use of average pooling in place of max pooling and progressively increasing dropout rates across layers. A schematic of the adapted CNN architecture is shown in Figure~\ref{urt_cnn}.

\subsubsection{Residual Networks (ResNets)}
\begin{figure}[!ht]
\centering
\includegraphics[width = 14.0 cm]{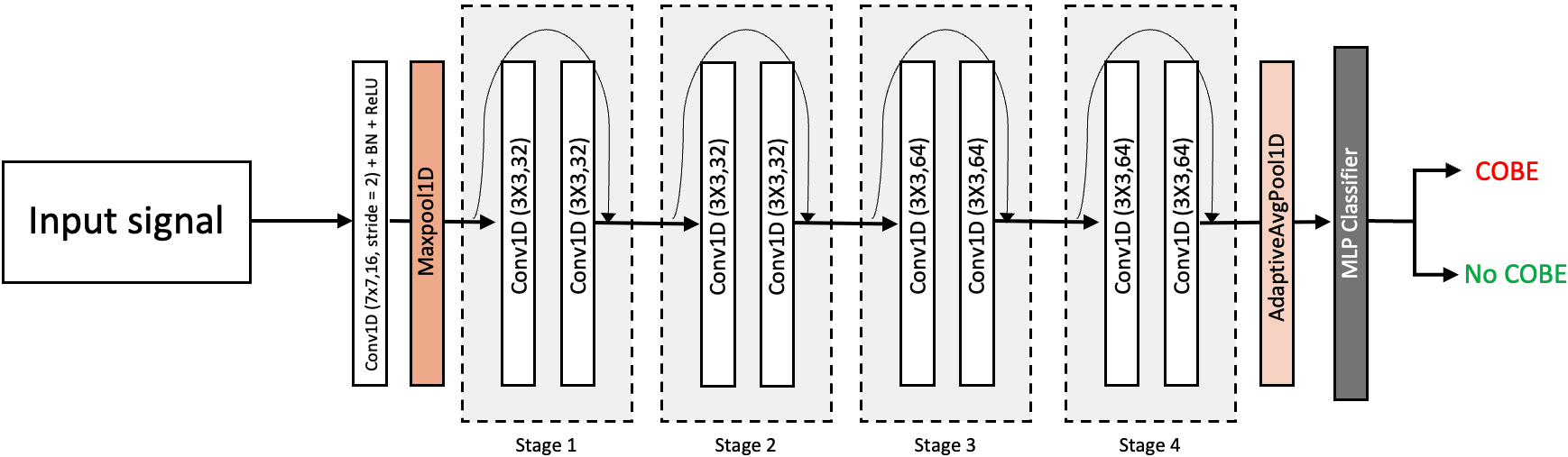}
\caption{The architecture of the ResNet model designed for automated COBE episode detection. The network consists of four sequential stages of residual blocks with multiple convolutional layers depending on the variant (ResNet-18, -34, or -50). The model is applied to individual input signals (IP, EDR, and PPG envelope) and extended to multimodal configurations using fusion strategies.}
\label{resnet_designed}
\end{figure}

ResNets were adapted for 1D physiological time series by replacing all 2D convolutional, pooling, and normalisation layers with their 1D counterparts. Following prior 1D adaptations~\cite{carter2023deep}, the number of feature channels per stage was reduced from [64, 128, 256, 512] to [32, 32, 64, 64], reflecting the reduced dimensionality of the input signals compared with 2D image data. Adaptive average pooling was used prior to classification to retain information from different regions of the feature map. The final classifier was extended to a multi-layer perceptron (MLP) to model non-linear relationships between pooled features and output labels. We evaluated ResNet-18, ResNet-34, and ResNet-50 architectures to assess the influence of network depth on COBE detection performance. The architecture is illustrated in figure~\ref{resnet_designed}.

\subsubsection{ConvNeXt architecture}
\begin{figure}[!ht]
\centering
\includegraphics[width = 14.0 cm]{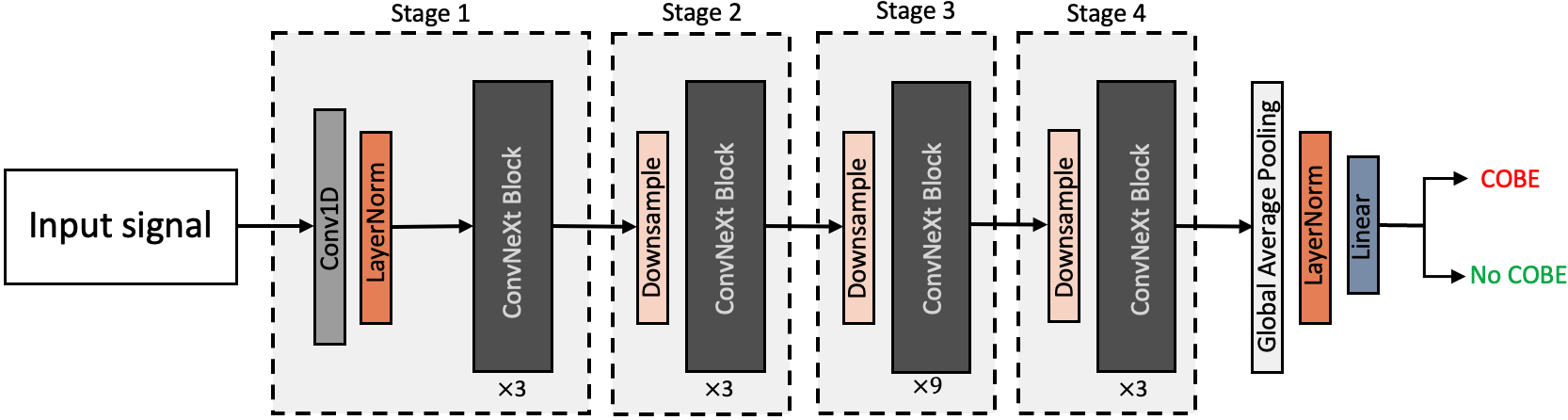}
\caption{Architecture of the modified 1D ConvNeXt model for automated COBE episode detection. The network processes 1D physiological signals (IP, EDR, and PPG envelope) through an initial 1D convolution and LayerNorm (commonly referred to as the ``stem''), followed by four stages of ConvNeXt blocks with intermediate downsampling. The final stage output is aggregated via global average pooling, normalised with LayerNorm, and classified with a linear layer into COBE or No COBE.}
           \label{conv_model}
\end{figure}

ConvNeXt, originally developed for 2D image classification, was adapted for 1D physiological time-series by replacing 2D convolutions, pooling operations, and normalisation layers with their 1D counterparts~\cite{talukder2024hybrid,zhu2024improved}.  The architecture processed temporal physiological signals directly rather than image representations. We evaluated both the default ConvNeXt kernel size (7) and a larger-kernel variant (21) to examine the influence of receptive field size on COBE detection performance, and also investigated reduced feature widths. Core architectural components of ConvNeXt, including Gaussian Error Linear Unit (GELU) activations, inverted bottlenecks, residual connections, and stage-wise downsampling, were retained. The resulting architecture processes each input signal through a stem layer followed by four stages of ConvNeXt blocks, with global average pooling and a linear classifier for final prediction. Multimodal configurations were implemented using early and late fusion strategies. The architecture is shown in Figure~\ref{conv_model}.
 
\subsection{Network training and evaluation protocol}
\label{train_prtcl}
The labelled dataset comprised 6,678 20-second windows derived from 954 annotated segments.  Each 20-second window was treated as a training sample. For model development and evaluation, the dataset was divided into six subsets, each containing data from four infants. The subsets were constructed to achieve a similar distribution of gestational age, sex, and class. One subset (16.7\% of the data) was reserved as an independent test set. The remaining five subsets were used for five-fold cross-validation. During cross-validation, models were trained on four subsets and validated on the remaining subset, with each subset serving as the validation set once. To prevent data leakage, infant-level separation was maintained throughout cross-validation and test evaluation, such that data from the same infant never appeared in both training and validation or test sets. After model selection, the final configuration was retrained using all five cross-validation subsets and evaluated on the independent test set.

\paragraph{}
Class imbalance was addressed using class-weighted loss functions. CNN and ResNet models were trained using the Adam optimiser for up to 100 epochs, with early stopping applied when validation loss failed to improve for five consecutive epochs. Learning rates were sampled logarithmically between $9 \times 10^{-5}$ and $1 \times 10^{-4}$, with exponential decay applied at each epoch. ConvNeXt models were trained using the AdamW optimiser, following the optimisation strategy proposed in the original ConvNeXt architecture~\cite{Liu_2022_CVPR}. A cosine learning-rate schedule with a five-epoch warm-up period was applied. Additional regularisation included weight decay (0.05), label smoothing (0.1), and exponential moving averages of model weights, consistent with standard ConvNeXt training practice~\cite{Liu_2022_CVPR}.

\paragraph{}
Performance was evaluated using balanced accuracy, sensitivity (true positive rate; TPR), precision, F1 score, and Cohen's $\kappa$. Cross-validation results were reported as mean $\pm$ standard deviation across folds. Model selection prioritised configurations with the highest mean cross-validation balanced accuracy while favouring lower variance when performance was comparable. Final performance was reported on the independent test set. Balanced accuracy was used to account for class imbalance by averaging sensitivity across COBE and normal breathing classes, given by:

\begin{equation}
\label{bal_accuracy}
Balanced \; Accuracy = \frac{{TPR}_{cobe} + {TPR}_{no\;cobe}}{2}\times 100\text{\%}
\end{equation}

\section{Results}

\label{cnn_best}
Table~\ref{summary_cnn} summarises CNN performance for unimodal and multimodal input configurations on the independent test set. The IP-only CNN achieved balanced accuracy of 86.8\% on the independent test set (TPR = 0.91), and the IP+PPG late-fusion CNN achieved the highest CNN F1 score (0.73). Full cross-validation results are provided in the Supplementary results (tables~\ref{ip_crossval}--\ref{multi_cnn_testLF}).

\begin{table}[!ht]
\centering
\caption{Performance of CNN baseline models for single- and multi-signal configurations on the test set. Best-performing configurations within each category are shown in bold.}
\label{summary_cnn}
\scalebox{0.85}{
\begin{tabular}{|>{\raggedright\arraybackslash}p{2.9cm}|R{1.5cm}|R{1.5cm}|R{1.5cm}|R{1.5cm}|R{1.5cm}|R{2cm}|R{2cm}|R{2cm}|}
\hline
\multirow{4}{*}{\parbox[c]{2.9cm}{\centering Signal(s)}} & \multicolumn{8}{c|}{Performance Metrics} \\
\cline{2-9}
 & \multicolumn{1}{c|}{TPR $\uparrow$} & \multicolumn{1}{c|}{FPR $\downarrow$} & \multicolumn{1}{c|}{Precision} & \multicolumn{1}{c|}{F1 Score} & \multicolumn{1}{c|}{\parbox{1.3cm}{\centering Cohen\\Kappa}} & \multicolumn{1}{c|}{\parbox{2cm}{\centering No COBE\\Accuracy (\%)}} & \multicolumn{1}{c|}{\parbox{2cm}{\centering COBE\\Accuracy (\%)}} & \multicolumn{1}{c|}{\parbox{2cm}{\centering Balanced\\Accuracy (\%)}} \\
 \hline
\multicolumn{9}{c}{\textbf{\textcolor{orange}{Individual signal training}}} \\ 
\hline
\textbf{IP} & \textbf{0.91} & \textbf{0.18} & \textbf{0.60} & \textbf{0.72} & \textbf{0.62 }& \textbf{82.5} & \textbf{91.1} & \textbf{86.8} \\
\hline
$ECG_{edr}$ & 0.64 & 0.38 & 0.32 & 0.43 & 0.19 & 61.6 & 63.6 & 62.6 \\
\hline
$PPG_{env}$ & 0.61 & 0.30 & 0.36 & 0.46 & 0.24 & 69.6 & 60.7 & 65.1 \\
\hline 
\multicolumn{9}{c}{\textbf{\textcolor{orange}{Multi-signal training: Late fusion }}} \\ 
\hline
\multirow{1}{*}{\parbox{1.6cm}{IP + ECG}} & 0.89 & 0.17 & 0.60 & 0.72 & 0.62 & 83.2 & 89.5 & 86.3 \\
\hline
\multirow{1}{*}{\parbox{1.7cm}{\textbf{IP + PPG}}}  & \textbf{0.88 }& \textbf{0.15} & \textbf{0.63} & \textbf{0.73} & \textbf{0.64} & \textbf{85.2} & \textbf{87.9} & \textbf{86.6} \\
\hline
\multirow{1}{*}{\parbox{1.9cm}{ECG  + PPG}} & 0.64 & 0.26 & 0.41 & 0.50 & 0.32 & 73.7 & 64.2 & 69.0 \\
\hline
\multirow{1}{*}{\parbox{2.5cm}{ IP + ECG  + PPG}} & 0.88 & 0.18 & 0.58 & 0.70 & 0.59 & 81.7 & 88.5 & 85.1 \\
\hline
  \multicolumn{8}{l}{\footnotesize $\uparrow$ indicates higher values are better; $\downarrow$ indicates lower values are better.} \\

\end{tabular}}
\end{table}

\label{resnet_best}
Table~\ref{summary_resnet} reports performance of the best ResNet variants on the independent test set. The IP-based ResNet-34 achieved balanced accuracy of 87.7\% on the independent test set (TPR = 0.93).  Supplementary results, including cross-validation metrics and performance across all ResNet depths, are provided in tables~\ref{resnet_cv}--\ref{multi_res_testEF}.

\begin{table}[!ht]
\centering
\caption{Performance of the ResNet variants for single- and multi-signal inputs on the independent test set. For each signal or signal combination, the ResNet depth yielding the highest validation performance is reported. Best-performing configurations are shown in bold.}
\label{summary_resnet}
\scalebox{0.80}{
\begin{tabular}{|>{\raggedright\arraybackslash}p{2.9cm}|>{\raggedright\arraybackslash}p{1.2cm}|R{1.5cm}|R{1.5cm}|R{1.5cm}|R{1.5cm}|R{1.9cm}|R{2cm}|R{2cm}|R{2.3cm}|}
\hline
\multirow{4}{*}{\parbox[c]{2.9cm}{\centering Signal(s)}}&\multirow{4}{*}{\parbox[c]{1.2cm}{\centering Resnet Depth}} & \multicolumn{8}{c|}{Performance Metrics} \\
\cline{3-10}
& & \multicolumn{1}{c|}{TPR$\uparrow$} & \multicolumn{1}{c|}{FPR$\downarrow$} & \multicolumn{1}{c|}{Precision} &
\multicolumn{1}{c|}{F1 Score} & \multicolumn{1}{c|}{\parbox{1.3cm}{\centering Cohen Kappa}} & \multicolumn{1}{c|}{\parbox{2cm}{\centering No COBE\\Accuracy (\%)}} & \multicolumn{1}{c|}{\parbox{2cm}{\centering COBE\\Accuracy (\%)}} & \multicolumn{1}{c|}{\parbox{2cm}{\centering Balanced\\Accuracy (\%)}} \\
\hline
\multicolumn{10}{c}{\textbf{\textcolor{orange}{Individual signal training}}} \\ 
\hline
\textbf{IP}                     & \textbf{34} & \textbf{0.93} & \textbf{0.17} & \textbf{0.61 }& \textbf{0.73} & \textbf{0.63} & \textbf{82.8} & \textbf{92.7} & \textbf{87.7} \\
\hline
$ECG_{edr}$ & 50 & 0.70 & 0.34 & 0.37 & 0.49 & 0.28 & 66.1 & 70.3 & 68.2 \\
\hline
$PPG_{env}$ & 34 & 0.63 & 0.31 & 0.37 & 0.47 & 0.26 & 69.5 & 62.6 & 66.1 \\
\hline
\multicolumn{10}{c}{\textbf{\textcolor{orange}{Multi-signal training: Late fusion}}} \\ 
\hline
\multirow{1}{*}{\parbox{1.5cm}{IP + ECG}} & - & 0.92 & 0.20 & 0.56 & 0.70 & 0.58 & 79.6 & 92.0 & 85.8 \\
\hline
\multirow{1}{*}{\parbox{1.6cm}{\textbf{IP + PPG}}} & - & \textbf{0.91} & \textbf{0.17} & \textbf{0.60} & \textbf{0.73} & \textbf{0.63} & \textbf{82.8} & \textbf{91.4} & \textbf{87.1 }\\
\hline
\multirow{1}{*}{\parbox{1.9cm}{ECG + PPG}} & - & 0.50 & 0.18 & 0.44 & 0.47 & 0.30 & 81.9 & 49.8 & 65.9 \\
\hline
\multirow{1}{*}{\parbox{2.5cm}{IP + ECG + PPG}} & - & 0.91 & 0.18 & 0.59 & 0.71 & 0.61 & 81.6 & 91.1 & 86.3 \\
\hline
\multicolumn{10}{c}{\textbf{\textcolor{orange}{Multi-signal training: Early fusion}}} \\
\hline
\multirow{1}{*}{\parbox{2.5cm}{IP + ECG + PPG}} & 34 & 0.91 & 0.18 & 0.59 & 0.72 & 0.62 & 82.0 & 91.4 & 86.7 \\
\hline
  \multicolumn{8}{l}{\footnotesize $\uparrow$ indicates higher values are better; $\downarrow$ indicates lower values are better.} \\
\end{tabular}}
\end{table}

Table~\ref{summary_convnext} summarises ConvNeXt performance on the independent test set.  
The IP-only ConvNeXt (larger kernels) achieved balanced accuracy of 88.0\% on the independent test set (TPR = 0.91), and the IP+PPG late-fusion ConvNeXt obtained the highest balanced test accuracy (88.7\%) and F1 (0.75). Full ConvNeXt results are provided in the Supplementary results (tables~\ref{conv_cv}--\ref{multi_conv_testEF}).

\begin{table}[!ht]
\centering
\caption{Performance of ConvNeXt architectures for single- and multi-signal configurations on the independent test set. For each input modality, the ConvNeXt variant achieving the highest validation performance is shown. Best-performing configurations are highlighted in bold.}
\label{summary_convnext}
\scalebox{0.75}{
\begin{tabular}{|>{\raggedright\arraybackslash}p{2.7cm}|>{\raggedright\arraybackslash}p{2.5cm}|R{1.5cm}|R{1.5cm}|R{1.5cm}|R{1.5cm}|R{1.9cm}|R{2cm}|R{2cm}|R{2.3cm}|}

\hline
\multirow{4}{*}{\parbox[c]{2.9cm}{\centering Signal(s)}}&\multirow{4}{*}{\parbox[c]{2.5cm}{\centering ConvNeXt variant}} & \multicolumn{8}{c|}{Performance Metrics} \\
\cline{3-10}
& & \multicolumn{1}{c|}{TPR$\uparrow$} & \multicolumn{1}{c|}{FPR$\downarrow$} & \multicolumn{1}{c|}{Precision} &
\multicolumn{1}{c|}{F1 Score} & \multicolumn{1}{c|}{\parbox{1.3cm}{\centering Cohen Kappa}} & \multicolumn{1}{c|}{\parbox{2cm}{\centering No COBE\\Accuracy (\%)}} & \multicolumn{1}{c|}{\parbox{2cm}{\centering COBE\\Accuracy (\%)}} & \multicolumn{1}{c|}{\parbox{2cm}{\centering Balanced\\Accuracy (\%)}} \\
\hline
\multicolumn{10}{c}{\textbf{\textcolor{orange}{Individual signal training}}} \\ 
\hline
\multirow{2}{*}{\textbf{IP}} & \textbf{Larger Kernels} &\multirow{2}{*}{ \textbf{0.91}} & \multirow{2}{*}{\textbf{0.15}} & \multirow{2}{*}{\textbf{0.64}} & \multirow{2}{*}{\textbf{0.75}} & \multirow{2}{*}{\textbf{0.66}} & \multirow{2}{*}{\textbf{85.3 }} & \multirow{2}{*}{\textbf{90.7}} & \multirow{2}{*}{\textbf{88.0}} \\
\hline
$ECG_{edr}$ & Default & 0.77 & 0.37 & 0.37 & 0.50 & 0.29 & 62.7 & 76.7 & 69.7 \\
\hline
$PPG_{env}$ & Default & 0.73 & 0.40 & 0.34 & 0.47 & 0.23 & 60.0 & 72.8 & 66.4 \\
\hline
\multicolumn{10}{c}{\textbf{\textcolor{orange}{Multi-signal training: Late fusion}}} \\ 
\hline
\multirow{1}{*}{\parbox{1.5cm}{IP + ECG}} & - & 0.91 & 0.17 & 0.61 & 0.73 & 0.63 & 83.0 & 90.7 & 86.9 \\
\hline
\multirow{1}{*}{\parbox{1.6cm}{\textbf{IP + PPG}}} & - & \textbf{0.94} & \textbf{0.16} & \textbf{0.62} &\textbf{ 0.75} & \textbf{0.66} & \textbf{83.9} & \textbf{93.6} & \textbf{88.7} \\
\hline
\multirow{1}{*}{\parbox{1.9cm}{ECG + PPG}} & - & 0.80 & 0.37 & 0.38 & 0.51 & 0.31 & 62.9 & 79.6 & 71.2 \\
\hline
\multirow{1}{*}{\parbox{2.5cm}{IP + ECG + PPG}} & - & 0.92 & 0.16 & 0.62 & 0.74 & 0.65 & 84.1 & 92.3 & 88.2 \\
\hline
\multicolumn{10}{c}{\textbf{\textcolor{orange}{Multi-signal training: Early fusion}}} \\
\hline
\multirow{1}{*}{\parbox{2.5cm}{IP + ECG + PPG}} & Default & 0.88 & 0.14 & 0.64 & 0.74 & 0.65 & 85.6 & 87.9 & 86.7 \\
\hline
\multirow{1}{*}{\parbox{2.5cm}{IP + ECG + PPG}} & Larger Kernels & 0.93 & 0.19 & 0.59 & 0.72 & 0.61 & 81.3 & 92.7 & 87.0 \\
\hline
\multirow{2}{*}{\parbox{2.5cm}{IP + ECG + PPG}} & Reduced Features & \multirow{2}{*}{0.90} & \multirow{2}{*}{0.18} & \multirow{2}{*}{0.59} & \multirow{2}{*}{0.71} & \multirow{2}{*}{0.61} & \multirow{2}{*}{81.8} & \multirow{2}{*}{90.4} & \multirow{2}{*}{86.1} \\
\hline
\multicolumn{8}{l}{\footnotesize $\uparrow$ indicates higher values are better; $\downarrow$ indicates lower values are better.} \\
\end{tabular}}
\end{table}

\section{Discussion}
This study evaluated the detection of apnoea-related COBE in pre-term infants using routinely acquired physiological signals (IP, ECG, and PPG) in the NICU. Although these signals are widely used in clinical monitoring and adult apnoea research, their systematic evaluation using deep learning approaches for neonatal COBE detection has been limited, largely due to challenges in collecting and annotating physiological data from pre-term infant populations. By comparing a CNN,  ResNet variants, and a 1D ConvNeXt architecture under similar preprocessing, windowing, and cross-validation protocols, this work provides a controlled assessment of how architectural complexity and signal modality influence detection performance.

\paragraph{}
The best-performing configuration (ConvNeXt, IP+PPG) achieved balanced accuracy  of 88.7\% on the independent test set, with an F1 score of 0.75 and the highest sensitivity among the evaluated models. Although improvements across architectures were modest, the results suggest that deep learning models can integrate complementary physiological information to improve discrimination of COBE events. 

\paragraph{}
Across all evaluated architectures, IP-based models consistently achieved higher balanced accuracy than ECG-derived and PPG-derived respiratory signals, suggesting that signal modality had a larger influence on detection performance than architectural choice. Unimodal IP-based models achieved test accuracies ranging from 86.8\% to 88.0\%, consistently outperforming ECG-derived (62.6-69.7\%) and PPG-derived (65.1-66.4\%) respiratory surrogates. This finding is consistent with IP providing a direct measure of respiratory effort, whereas ECG- and PPG-derived signals represent indirect respiratory surrogates. However, these findings should be interpreted in the context of the limited cohort size (24 infants) and the specific preprocessing pipeline used. Multimodal fusion produced incremental gains in some configurations, most notably for the IP+PPG ConvNeXt model, but improvements were modest relative to IP alone. This trend was consistent across CNN, ResNet, and ConvNeXt architectures, indicating that architectural depth plays a secondary role to the discriminative properties of the underlying physiological signal.

Table~\ref{best_models_summary} summarises the highest-performing configurations across architectures and input modalities, highlighting that IP-based models consistently achieve higher performance than ECG- and PPG-derived inputs, with only modest gains from multimodal fusion. Detailed results for all model configurations and per-fold statistics are provided in the Supplementary Material (Appendices~\ref{appendix:supp_cnn} -- \ref{appendix:supp_convnext}).

\begin{table}[!ht]
\centering
\caption{Summary of highest-performing models across architectures and input modalities, reporting mean cross-validation balanced accuracy (mean $\pm$ standard deviation) and balanced accuracy on the independent test set. Architectures and configuration variants correspond to those achieving the best validation performance within each category.}
\label{best_models_summary}
\scalebox{0.75}{
\begin{tabular}{|>{\raggedright\arraybackslash}p{3.0cm} |>{\raggedright\arraybackslash}p{3.5cm}|>{\raggedright\arraybackslash}p{3.0cm}|>{\raggedright\arraybackslash}p{4.5cm} |R{2.5cm}|R{2.5cm}|}
\hline
\multirow{2}{*}{\parbox[c]{2.9cm}{\centering Input modality}}&\multirow{2}{*}{\parbox[c]{3.5cm}{\centering Input Signal(s)}} &\multirow{2}{*}{\parbox[c]{2.9cm}{\centering Architecture}} & \multirow{2}{*}{\parbox[c]{4.5cm}{\centering Modification}} & \multicolumn{2}{c|}{Balanced Accuracy (\%)} \\
\cline{5-6}
& & & & \multicolumn{1}{c|}{Validation} & \multicolumn{1}{c|}{Test} \\
\hline
\hline
\multirow{5}{*}{\parbox{1.8cm}{Individual}} & IP & ConvNeXt & Larger Kernels & 86.7 $\pm$ 1.9 & 88.0\\
\cline{2-6}
 & \multirow{1}{*}{\parbox{0.9cm}{$ECG_{edr}$}} & ConvNeXt & default & 71.9 $\pm$ 5.5 & 69.7\\
 \cline{2-6}
 & \multirow{1}{*}{\parbox{0.9cm}{$ECG_{raw}$}} & ResNet & ResNet-18 & 58.0 $\pm$  7.0 & 65.0  \\
 \cline{2-6}
 & \multirow{1}{*}{\parbox{0.9cm}{$PPG_{env}$}} & ConvNeXt & default & 69.6 $\pm$ 3.6 & 66.4 \\
 \cline{2-6}
 & \multirow{1}{*}{\parbox{0.9cm}{$PPG_{raw}$}} & ResNet& ResNet-34 & 66.0 $\pm$ 3.3 & 65.8 \\
\hline
\hline
\multirow{7}{*}{\parbox{1.8cm}{Multimodal}} & \multirow{2}{*}{\parbox{1.5cm}{IP + ECG}} & \multirow{2}{*}{ConvNeXt} & Larger kernels (IP), default model (ECG) & \multirow{2}{*}{85.1 $\pm$ 3.2} & \multirow{2}{*}{86.9} \\
\cline{2-6}
 & \multirow{2}{*}{\parbox{1.4cm}{IP + PPG}} & \multirow{2}{*}{ConvNeXt} & Larger kernels (IP), default model (PPG) & \multirow{2}{*}{85.6 $\pm$ 3.3} & \multirow{2}{*}{88.7}\\
 \cline{2-6}
 & \multirow{1}{*}{\parbox{1.9cm}{ECG + PPG}} & ConvNeXt & default & 72.2 $\pm$ 4.8 & 71.2 \\
 \cline{2-6}
 & \multirow{2}{*}{\parbox{2.5cm}{IP + ECG + PPG}} & \multirow{2}{*}{ConvNeXt} & Larger kernels (IP), default model (PPG and ECG) & \multirow{2}{*}{85.4 $\pm$ 3.0} & \multirow{2}{*}{88.2}\\
\hline
\end{tabular}}
\end{table}

\newpage
\paragraph{}
It is important to acknowledge that $RR$ formed part of the annotation workflow used to define COBE events, introducing partial alignment between IP-derived features and ground truth labels. However, annotation decisions were not based solely on thresholded $RR$ values; reviewers considered concurrent physiological signals and the surrounding 5-minute recording window when assigning labels. The consistent performance differences between IP and both ECG- and PPG-derived inputs suggest that IP dominance reflects not only annotation alignment but also its more direct representation of respiratory mechanics. Future work incorporating independent airflow measurements would further clarify modality-specific contributions.

\paragraph{}
Architectural complexity influenced performance, but to a lesser extent than signal modality. ResNet models improved balanced accuracy on the independent test set relative to the shallow CNN (e.g., 87.7\% vs 86.8\% for IP-only inputs), and ConvNeXt achieved the highest overall performance (88.7\% for IP+PPG). However, gains over ResNet were modest (approximately 1--2 percentage points), suggesting that architectural refinements provide incremental improvements but do not fundamentally alter the signal-dependent nature of COBE detection in our dataset. Interpretation of these differences should consider that the ConvNeXt models were trained using a different optimisation strategy from the CNN and ResNet models. ConvNeXt employed AdamW optimisation, cosine learning-rate scheduling with warm-up, exponential moving averages, and label smoothing, whereas the CNN and ResNet models used Adam optimisation with exponential learning-rate decay. Consequently, the superior performance of ConvNeXt cannot be attributed solely to architectural differences.

\paragraph{}
Compared to prior neonatal apnoea detection studies, which often rely on single-signal inputs, handcrafted features, or heterogeneous segmentation strategies, this work provides a systematic comparison of architecture families and multimodal fusion under consistent preprocessing and evaluation protocols. Previous neonatal studies have reported accuracies approaching 87\% using smaller cohorts, different windowing strategies,  and evaluation protocols. In contrast, the present study evaluates multiple deep architectures under similar data preparation and validation procedures, enabling clearer assessment of how signal modality and architectural complexity influence performance. To our knowledge, this represents the first application of a 1D ConvNeXt architecture for apnoea-related breathing cessation detection in pre-term infants.

\paragraph{}
From a clinical perspective, bedside monitors generate apnoea alarms using routinely acquired respiratory and oxygenation signals. The present findings demonstrate that deep learning models applied to these same physiological inputs can achieve strong detection performance of COBE events without requiring additional sensing hardware. Prospective validation of alarm burden within operational workflows will be necessary before translation into clinical practice.

\section{Conclusion}
This study demonstrates that routinely acquired neonatal physiological signals can support automated detection of apnoea-related COBE using modern deep learning architectures. The best-performing ConvNeXt configuration achieved balanced accuracy of 88.7\% on the independent test set with an F1 score of 0.75.

\paragraph{}
Across models, detection performance was primarily influenced by the underlying physiological signal rather than architectural complexity. Impedance pneumography (IP) provided the most discriminative information, while ECG- and PPG-derived respiratory surrogates yielded lower balanced test accuracy. Multimodal fusion offered modest, architecture-dependent gains. To our knowledge, this is the first systematic evaluation of a 1D ConvNeXt architecture for neonatal COBE detection, alongside controlled comparisons with CNN and ResNet models under identical preprocessing and evaluation protocols. Although deep learning is well established in adult apnoea detection, its application in neonatal populations remains limited, and systematic comparisons across modern convolutional architectures have not been reported. These findings clarify the relative contributions of signal modality and architecture in data-constrained neonatal monitoring settings and highlight the potential of routinely available NICU signals for improving automated COBE detection, particularly for events associated with clinically significant oxygen desaturation.

\newpage
\appendix
\renewcommand{\thetable}{S\arabic{table}}
\setcounter{table}{0} 

\section{Supplemetary methods: Manual annotation of COBE events}
\label{appendix:supp_dataset}

\subsection{MATLAB annotation interface}
    \label{annot_gui}
The MATLAB-based graphical user interface (GUI) was designed to support manual review of synchronised physiological signals extracted from the Neonatal HDU dataset, including ECG, PPG, IP,  $RR$, monitor-derived heart rate, and $SpO_2$. Annotators used the interface to determine whether candidate segments satisfied the study definition of COBE, characterised by either (1) periods where $RR$ remained below 20 breaths/min for at least 20~s, or (2) shorter respiratory pauses lasting at least 10~s accompanied by oxygen desaturation ($SpO_2 \leq$ 80\%)~\cite{zhao2011apnea,bester2018study}. 

Primary annotation decisions were initially based on the $RR$ and $SpO_2$ signals. For difficult annotation cases or suspected signal artefacts, candidate segments could then be reloaded in an extended review mode containing supplementary ECG, PPG,  and raw IP waveforms to support annotation decisions. Figure~\ref{annot_gui} shows the MATLAB-based annotation interface used for this extended multi-signal review process. The interface supported zooming, signal rescaling, and detailed inspection of ambiguous segments containing possible motion artefacts or sensor disconnections.

\begin{figure}[!ht]
    \centering
    \includegraphics[width=15.5cm]{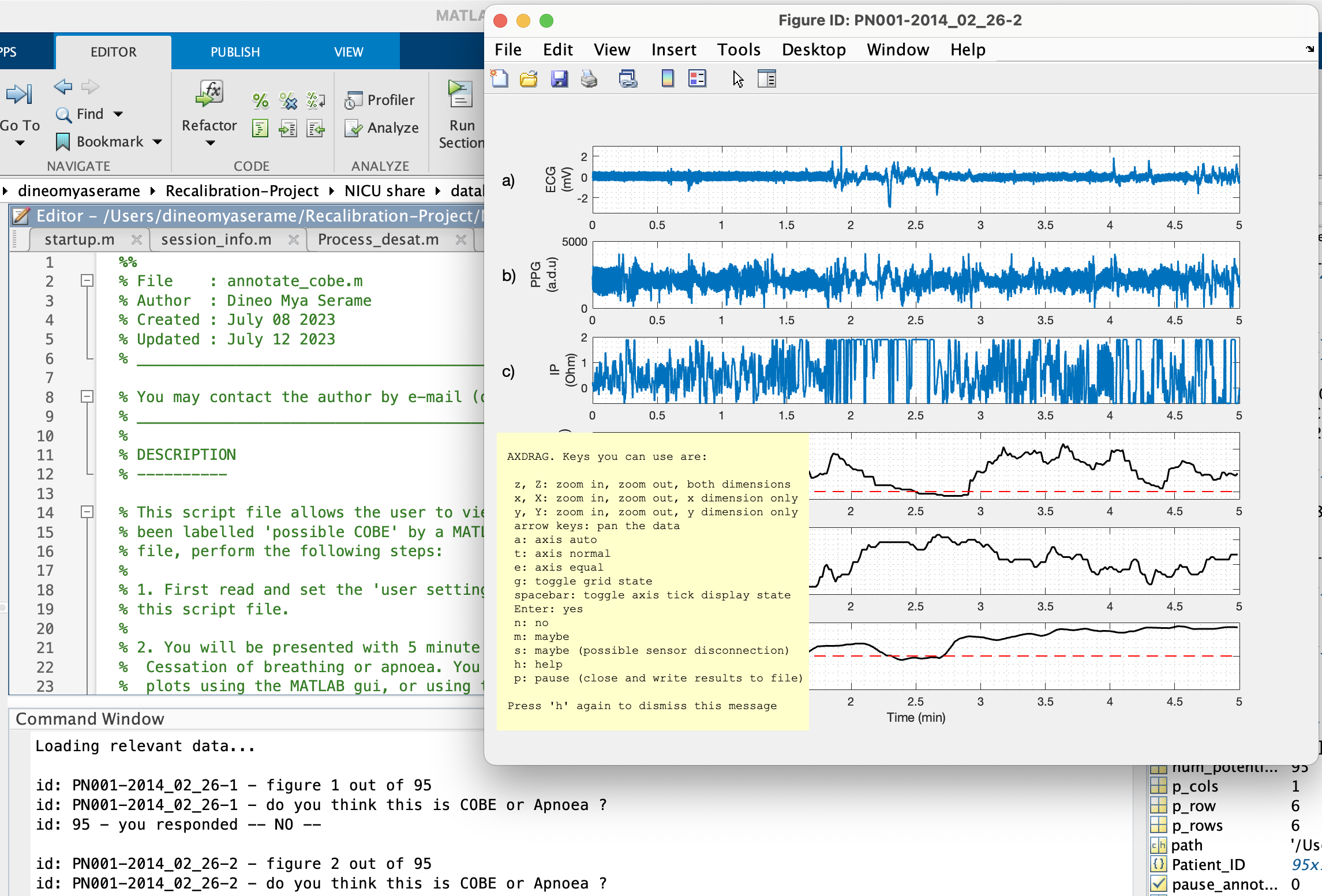}
    \vspace*{1mm}
    \caption[MATLAB annotation interface used for COBE dataset review.]{MATLAB annotation interface used for expert review of candidate COBE episodes. The GUI displays multiple synchronised physiological signals, including ECG, PPG, IP, $RR$, $HR$, and $SpO_2$, alongside annotation controls and keyboard shortcuts. The command window shows an example segment already annotated as ``NO''.}
\end{figure}

\subsection{Annotation decision workflow}
\label{annotate_process}
\begin{figure}[!ht]
\centering
\includegraphics[width = 12.5 cm]{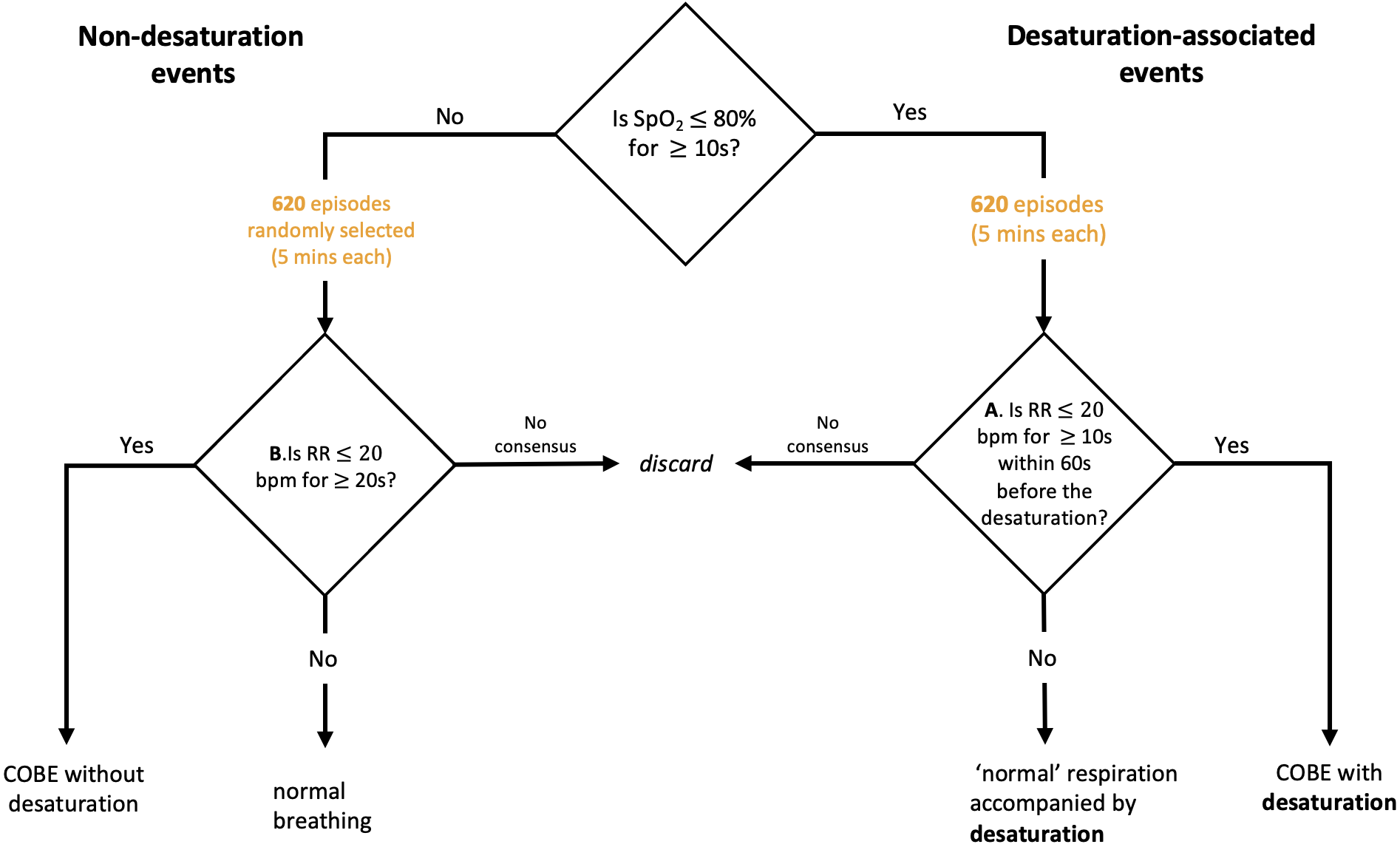}
\caption{Decision tree used for manual annotation of candidate COBE episodes. Reviewers assessed 5-minute windows of clinical vital signs surrounding each trigger event.}
\end{figure}

\section{Supplemetary methods: Respiratory signal extraction}
\label{appendix:supp_signals}

\subsection{Impedance pneumography (IP)}
The raw IP signal was filtered using an 8th-order high-pass Butterworth IIR filter (0.08 Hz) and a 6th-order low-pass Butterworth IIR filter (2.75 Hz) to remove baseline drift and high-frequency noise while preserving the neonatal respiratory frequency range. The filtered signal was then resampled to 60 Hz using cubic spline interpolation to ensure temporal alignment with other modalities. The resulting waveform was used directly as the respiratory input signal.

\begin{figure}[!ht]
\centering
\includegraphics[width=12.0cm]{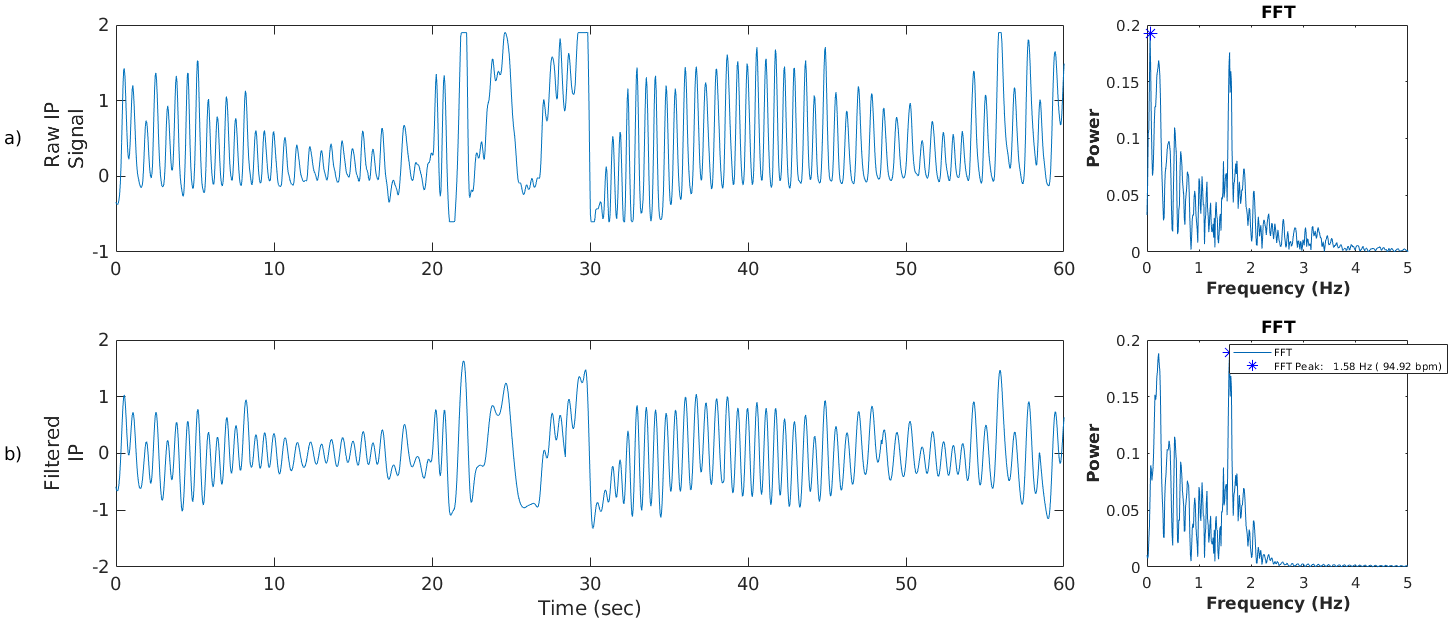}
\caption{IP waveform extracted from a 60-second IP segment acquired from a 28-week-old infant. a) Raw IP; b) Filtered IP waveform.}
\label{IP_sig}
\end{figure}

\subsection{ECG-derived respiration (EDR)}
The ECG signal was filtered using an 8th-order high-pass Butterworth IIR filter (0.67 Hz) and a 2nd-order low-pass Butterworth IIR filter (4 Hz) to preserve QRS complexes for reliable peak detection. R-peaks were identified using the Pan–Tompkins algorithm~\cite{pan1985real,chaichulee2018non}. Respiratory modulation was derived from variations in R-peak amplitude. Successive R-peak amplitudes were interpolated using cubic splines to generate a continuous ECG-derived respiration (EDR) waveform. The resulting signal captures low-frequency variations associated with respiration and was resampled to 60 Hz for alignment with other signals.

\begin{figure}[!ht]
\centering
\includegraphics[width=12.2cm]{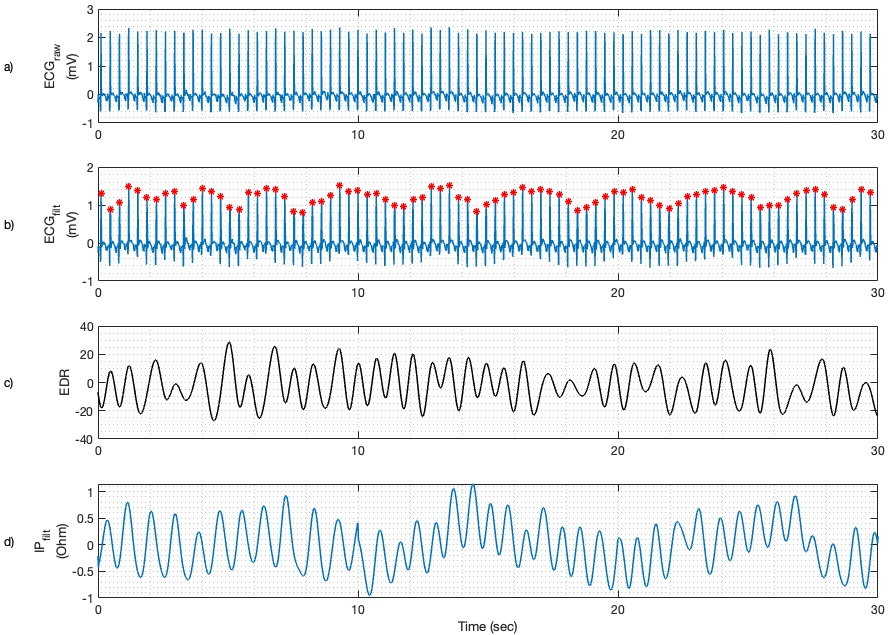}
\caption{EDR waveform extracted from a 60-second ECG segment acquired from a 30.3-week-old infant. a) Raw ECG; b) Filtered ECG waveform with detected R-peaks; c) EDR signal from successive R-peak amplitudes; d) IP signal during the same period.}
\label{edr_extr}
\end{figure}

\subsection{PPG-derived respiratory envelope}
Respiratory modulation in the PPG signal arises from changes in venous return and peripheral blood volume during breathing. The PPG signal was filtered using an 8th-order high-pass Butterworth IIR filter (0.08 Hz) and a 2nd-order low-pass Butterworth IIR filter (2.75 Hz) prior to extraction of respiratory-induced variations. The filtered signal was resampled to 60 Hz for temporal alignment. Respiratory information was extracted by detecting successive PPG peaks, which were interpolated using cubic splines to generate a continuous envelope signal. This envelope captures low-frequency amplitude variations associated with respiration and was used as the PPG-derived respiratory signal.

\begin{figure}[!ht]
\centering
\includegraphics[width=13.2cm]{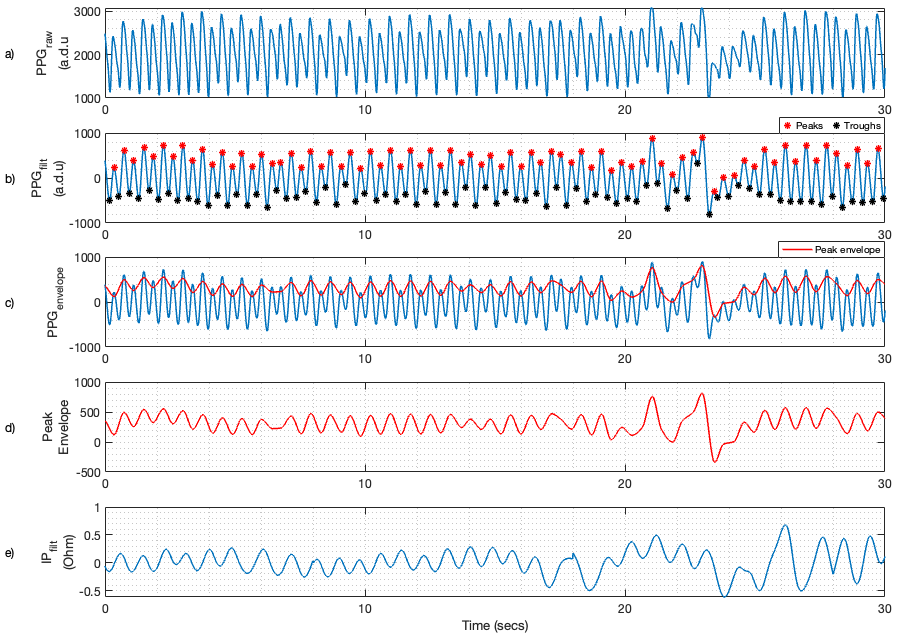}
\caption[PPG envelope waveform extraction from a 30-second segment]{PPG envelope waveform extracted from a 30-second PPG segment acquired from a 31.9-week-old infant. a) Raw PPG; b) Filtered PPG waveform with detected peaks and troughs; c) Envelope derived from peaks; d) Peak envelope representing respiratory variations; e) IP signal during the same period.}
\label{env_extr}
\end{figure}

\section{Supplemetary results: COBE detection using CNNs}
\label{appendix:supp_cnn}
This appendix reports cross-validation and test results for CNN experiments. Tables present final hyperparameters for selected models, cross-validation results for single-signal experiments, selected model test performance, and multi-signal late-fusion results. All models were trained using identical cross-validation splits and preprocessing pipelines as described in the main manuscript.

\subsection{Training using individual signals}
\subsubsection{IP signal}
Table~\ref{ip_crossval} reports five-fold cross-validation results (mean $\pm$ SD) for the evaluated CNN variants trained on the IP signal. The final IP model (Model B) is shown in Table~\ref{ip_train_test}.

\begin{table}[!ht]
\centering
\caption{Five-fold cross-validation results for CNN variants trained on IP. Metrics: TPR, FPR, Precision, F1, Cohen's Kappa, per-class accuracies and balanced accuracy.}
\label{ip_crossval}
\scalebox{0.75}{
\begin{tabular}{|>{\centering}p{2.0 cm}|>{\centering}p{2.0 cm}|>{\centering}p{2.0cm} |>{\centering}p{2.0cm}|>{\centering}p{2.0cm}|>{\centering}p{2.0cm}|>{\centering}p{2.1cm}|>{\centering}p{2.1cm}|P{2.1 cm}|}
\hline
\multirow{3}{*}{Model} & \multicolumn{8}{c|}{Performance Measure} \\
\cline{2-9}
 & TPR & FPR & Precision & F1 Score & Cohen Kappa & \textbf{No COBE class} accuracy (\%) & \textbf{COBE class} accuracy (\%) & Accuracy (\%) \\
\hline
\multicolumn{9}{c}{\textbf{\textcolor{orange}{Average Pooling Strategy}}} \\ 
\hline   
\multirow{1}{*}{\parbox{0.2cm}{A}}  & 0.88 $\pm$ 0.04 & 0.16 $\pm$ 0.03 & 0.48 $\pm$ 0.07 & 0.62 $\pm$ 0.05 & 0.52 $\pm$ 0.06 & 83.6 $\pm$ 3.6 & 88.3 $\pm$ 4.5 & 86.0 $\pm$ 2.4 \\ 
\hline
\rowcolor{green!15}
\multirow{1}{*}{\parbox{0.2 cm}{B}} & 0.90 $\pm$ 0.06 & 0.16 $\pm$ 0.04 & 0.48 $\pm$ 0.07 & 0.62 $\pm$ 0.06 & 0.54 $\pm$ 0.07 & 83.6 $\pm$ 4.4 & 90.0 $\pm$ 5.7 & 86.8 $\pm$ 3.7 \\
\hline
\multirow{1}{*}{\parbox{0.2 cm}{C}}  & 0.90 $\pm$ 0.07 & 0.19 $\pm$ 0.04 & 0.44 $\pm$ 0.10 & 0.58 $\pm$ 0.08 & 0.48 $\pm$ 0.08 & 80.8 $\pm$ 3.7 & 90.2 $\pm$ 7.1 & 85.5 $\pm$ 2.5 \\
\hline
\multirow{1}{*}{\parbox{0.2cm}{D}}  & 0.89 $\pm$ 0.05 & 0.17 $\pm$ 0.05 & 0.47 $\pm$ 0.07 & 0.61 $\pm$ 0.06 & 0.52 $\pm$ 0.06 & 82.6 $\pm$ 5.2 & 88.7 $\pm$ 5.0 & 85.6 $\pm$ 2.8 \\
\hline
\multicolumn{9}{c}{\textbf{\textcolor{orange}{Max Pooling Strategy}}}\\ 
\hline
\rowcolor{green!5}
\multirow{1}{*}{\parbox{0.75cm}{$A_{max}$}} & 0.89 $\pm$ 0.05 & 0.17 $\pm$ 0.02 & 0.46 $\pm$ 0.06 & 0.60 $\pm$ 0.06 & 0.51 $\pm$ 0.05 & 82.5 $\pm$ 2.2 & 89.3 $\pm$ 4.3 & 85.9 $\pm$ 2.4 \\
\hline
 \multirow{1}{*}{\parbox{0.75cm}{$B_{max}$}} & 0.91 $\pm$ 0.04 & 0.20 $\pm$ 0.06 & 0.44 $\pm$ 0.09 & 0.59 $\pm$ 0.08 & 0.49 $\pm$ 0.09 & 80.2 $\pm$ 5.9 & 91.6 $\pm$ 3.9 & 85.9 $\pm$ 1.9 \\ 
\hline
\multirow{1}{*}{\parbox{0.75cm}{$C_{max}$}}  & 0.87 $\pm$ 0.06 & 0.17 $\pm$ 0.05 & 0.47 $\pm$ 0.06 & 0.61 $\pm$ 0.05 & 0.51 $\pm$ 0.06 & 83.1 $\pm$ 4.8 & 87.4 $\pm$ 5.6 & 85.2 $\pm$ 3.0 \\
\hline
\multirow{1}{*}{\parbox{0.75cm}{$D_{max}$}} & 0.92 $\pm$ 0.03 & 0.21 $\pm$ 0.06 & 0.44 $\pm$ 0.06 & 0.59 $\pm$ 0.06 & 0.49 $\pm$ 0.07 & 79.6 $\pm$ 6.2 & 92.3 $\pm$ 3.2 & 85.9 $\pm$ 3.2 \\ 
\hline
\multicolumn{9}{c}{\textbf{\textcolor{orange}{Kernel Size change}}} \\ 
\hline
\multirow{1}{*}{\parbox{0.2cm}{E}} & 0.93 $\pm$ 0.02 & 0.22 $\pm$ 0.06 & 0.41 $\pm$ 0.05 & 0.57 $\pm$ 0.05 & 0.47 $\pm$ 0.06 & 77.7 $\pm$ 6.2 & 93.2 $\pm$ 1.9 & 85.5 $\pm$ 3.4 \\
\hline
\multirow{1}{*}{\parbox{0.2cm}{F}} & - & - & - & - & - & - & - &  \\   
\hline
\multicolumn{9}{c}{\textbf{\textcolor{orange}{Deeper 8-layer model}}} \\
\hline
\multirow{1}{*}{\parbox{0.2cm}{G}} & 0.91 $\pm$ 0.05 & 0.24 $\pm$ 0.04 & 0.39 $\pm$ 0.07 & 0.54 $\pm$ 0.06 & 0.43 $\pm$ 0.06 & 76.2 $\pm$ 3.5 & 90.7 $\pm$ 4.9 & 83.4 $\pm$ 3.0 \\
\hline    
\end{tabular}}
\end{table}

\begin{table}[!ht]
\centering
\caption{Independent test-set performance of the selected IP CNN (Model B). Metrics computed on held-out test infants.}
\label{ip_train_test}
\scalebox{0.75}{
\begin{tabular}{|>{\centering}p{2.0 cm}|>{\centering}p{2.0 cm}|>{\centering}p{2.0cm} |>{\centering}p{2.0cm}|>{\centering}p{2.0cm}|>{\centering}p{2.0cm}|>{\centering}p{2.1cm}|>{\centering}p{2.1cm}|P{2.1 cm}|}
\hline
\multirow{3}{*}{Model} & \multicolumn{8}{c|}{Performance Measure } \\
 \cline{2-9}
 &  TPR & FPR & Precision & F1 Score & Cohen Kappa & \textbf{No COBE class} accuracy (\%) & \textbf{COBE class} accuracy (\%) & Balanced Accuracy (\%) \\
\hline   
   \multirow{1}{*}{\parbox{0.2cm}{B}} & 0.91 & 0.18 & 0.60 & 0.72 & 0.62 & 82.5 & 91.1 & 86.8 \\ 
\hline 
 \end{tabular}}
\end{table}

\newpage
\subsubsection{EDR signal}
Table~\ref{ECG_crossval} reports cross-validation results for CNN variants trained on the EDR signal. The selected EDR model test performance is reported in Table~\ref{ECG_train_test}.

\begin{table}[!ht]
\centering
\caption{Five-fold cross-validation results for CNN variants trained on EDR.}
\label{ECG_crossval}
\scalebox{0.78}{
\begin{tabular}{|>{\centering}p{2.0 cm}|>{\centering}p{2.0 cm}|>{\centering}p{2.0cm} |>{\centering}p{2.0cm}|>{\centering}p{2.0cm}|>{\centering}p{2.0cm}|>{\centering}p{2.1cm}|>{\centering}p{2.1cm}|P{2.1 cm}|}
\hline
\multirow{3}{*}{Model} & \multicolumn{8}{c|}{Performance Measure } \\
 \cline{2-9}
 & TPR & FPR & Precision & F1 Score & Cohen Kappa & \textbf{No COBE class} accuracy (\%) & \textbf{COBE class} accuracy (\%) & Accuracy (\%) \\
 
 \hline
\multicolumn{9}{c}{\textbf{\textcolor{orange}{Average Pooling Strategy}}} \\
\hline   
\multirow{1}{*}{\parbox{0.2cm}{A}}  & 0.68 $\pm$ 0.12 & 0.32 $\pm$ 0.09 & 0.26 $\pm$ 0.05 & 0.37 $\pm$ 0.06 & 0.21 $\pm$ 0.06 & 67.6 $\pm$ 8.8 & 68.1 $\pm$ 11.7 & 67.8 $\pm$ 5.3 \\
\hline
\rowcolor{green!15}
\multirow{1}{*}{\parbox{0.2 cm}{B}} & 0.68 $\pm$ 0.10 & 0.30 $\pm$ 0.04 & 0.28 $\pm$ 0.05 & 0.39 $\pm$ 0.06 & 0.24 $\pm$ 0.04 & 70.4 $\pm$ 3.9 & 68.5 $\pm$ 9.9 & 69.5 $\pm$ 3.7 \\ 
\hline
\multirow{1}{*}{\parbox{0.2 cm}{C}}  & 0.66 $\pm$ 0.12 & 0.32 $\pm$ 0.07 & 0.26 $\pm$ 0.06 & 0.37 $\pm$ 0.06 & 0.21 $\pm$ 0.04 & 68.5 $\pm$ 7.5 & 66.6 $\pm$ 12.2 & 67.6 $\pm$ 3.5 \\
\hline
\multirow{1}{*}{\parbox{0.2cm}{D}}  & 0.71 $\pm$ 0.05 & 0.33 $\pm$ 0.09 & 0.27 $\pm$ 0.07 & 0.39 $\pm$ 0.08 & 0.22 $\pm$ 0.08 & 66.4 $\pm$ 8.5 & 71.2 $\pm$ 4.9 & 68.8 $\pm$ 4.6 \\ 
\hline
\multicolumn{9}{c}{\textbf{\textcolor{orange}{Max Pooling Strategy}}} \\ 
\hline
\multirow{1}{*}{\parbox{0.9cm}{$A_{max}$}} & 0.66 $\pm$ 0.10 & 0.29 $\pm$ 0.08 & 0.28 $\pm$ 0.07 & 0.39 $\pm$ 0.07 & 0.24 $\pm$ 0.08 & 71.3 $\pm$ 8.3 & 65.5 $\pm$ 10.5 & 68.4 $\pm$ 6.1 \\
\hline
\rowcolor{green!5}
\multirow{1}{*}{\parbox{0.9cm}{$B_{max}$}} & 0.69 $\pm$ 0.12 & 0.30 $\pm$ 0.08 & 0.28 $\pm$ 0.04 & 0.40 $\pm$ 0.06 & 0.25 $\pm$ 0.05 & 70.4 $\pm$ 8.0 & 69.2 $\pm$ 11.8 & 69.8 $\pm$ 4.4 \\
\hline
\multirow{1}{*}{\parbox{0.9cm}{$C_{max}$}}  & 0.63 $\pm$ 0.05 & 0.28 $\pm$ 0.08 & 0.28 $\pm$ 0.08 & 0.38 $\pm$ 0.08 & 0.23 $\pm$ 0.08 & 71.5 $\pm$ 8.3 & 63.7 $\pm$ 5.0 & 67.6 $\pm$ 4.3 \\ 
\hline
\multirow{1}{*}{\parbox{0.9cm}{$D_{max}$}} & 0.66 $\pm$ 0.07 & 0.30 $\pm$ 0.14 & 0.29 $\pm$ 0.08 & 0.39 $\pm$ 0.08 & 0.24 $\pm$ 0.09 & 70.4 $\pm$ 14.1 & 66.4 $\pm$ 6.8 & 68.4 $\pm$ 4.8 \\ 
\hline
\multicolumn{9}{c}{\textbf{\textcolor{orange}{Kernel Size change}}} \\ 
\hline
\multirow{1}{*}{\parbox{0.2cm}{E}} & 0.70 $\pm$ 0.10 & 0.31 $\pm$ 0.04 & 0.27 $\pm$ 0.07 & 0.39 $\pm$ 0.08 & 0.24 $\pm$ 0.08 & 68.7 $\pm$ 4.2 & 70.4 $\pm$ 9.6 & 69.5 $\pm$ 5.6 \\ 
\hline
\multirow{1}{*}{\parbox{0.2cm}{F}} & - & - & - & - & - & - & - &  \\ 
\hline
\multicolumn{9}{c}{\textbf{\textcolor{orange}{Deeper 8-layer model}}} \\
\hline
\multirow{1}{*}{\parbox{0.2cm}{G}} & 0.67 $\pm$ 0.07 & 0.30 $\pm$ 0.03 & 0.27 $\pm$ 0.07 & 0.38 $\pm$ 0.07 & 0.23 $\pm$ 0.06 & 70.1 $\pm$ 2.9 & 67.3 $\pm$ 7.1 & 68.7 $\pm$ 4.1 \\
\hline 
\end{tabular}}
\end{table}

\begin{table}[!ht]
\centering
\caption{Independent test-set performance of the selected EDR CNN (Model B).}
\label{ECG_train_test}
\scalebox{0.78}{
\begin{tabular}{|>{\centering}p{2.0 cm}|>{\centering}p{2.0 cm}|>{\centering}p{2.0cm} |>{\centering}p{2.0cm}|>{\centering}p{2.0cm}|>{\centering}p{2.0cm}|>{\centering}p{2.1cm}|>{\centering}p{2.1cm}|P{2.1 cm}|}
\hline
\multirow{3}{*}{Model} & \multicolumn{8}{c|}{Performance Measure } \\
\cline{2-9}
 & TPR & FPR & Precision & F1 Score & Cohen Kappa & \textbf{No COBE class} accuracy (\%) & \textbf{COBE class} accuracy (\%) & Accuracy (\%) \\
\hline   
\multirow{1}{*}{\parbox{0.2cm}{B}} & 0.64 & 0.38 & 0.32 & 0.43 & 0.19 & 61.6 & 63.6 & 62.6 \\
\hline
 \end{tabular}}
\end{table}

\subsubsection{PPG envelope signal}
Table~\ref{PPG_crossval} reports cross-validation results for CNN variants trained on the PPG envelope. The selected PPG model test performance is reported in Table~\ref{PPG_train_test}.

\begin{table}[!ht]
\centering
\caption{Five-fold cross-validation results for CNN variants trained on the PPG envelope.}
\label{PPG_crossval}
\scalebox{0.78}{
\begin{tabular}{|>{\centering}p{2.0 cm}|>{\centering}p{2.0 cm}|>{\centering}p{2.0cm} |>{\centering}p{2.0cm}|>{\centering}p{2.0cm}|>{\centering}p{2.0cm}|>{\centering}p{2.1cm}|>{\centering}p{2.1cm}|P{2.1 cm}|}
\hline
\multirow{3}{*}{Model} & \multicolumn{8}{c|}{Performance Measure } \\
 \cline{2-9}
 & TPR & FPR & Precision & F1 Score & Cohen Kappa & \textbf{No COBE class} accuracy (\%) & \textbf{COBE class} accuracy (\%) & Accuracy (\%) \\
 \hline
 \multicolumn{9}{c}{\textbf{\textcolor{orange}{Average Pooling Strategy}}} \\ 
\hline   
\rowcolor{green!5}
\multirow{1}{*}{\parbox{0.2cm}{A}}  & 0.69 $\pm$ 0.07 & 0.37 $\pm$ 0.06 & 0.24 $\pm$ 0.07 & 0.35 $\pm$ 0.07 & 0.18 $\pm$ 0.06 & 62.8 $\pm$ 6.3 & 68.9 $\pm$ 7.0 & 65.8 $\pm$ 3.5 \\ 
\hline
\multirow{1}{*}{\parbox{0.2 cm}{B}} & 0.72 $\pm$ 0.12 & 0.39 $\pm$ 0.04 & 0.24 $\pm$ 0.06 & 0.35 $\pm$ 0.07 & 0.18 $\pm$ 0.05 & 61.4 $\pm$ 4.1 & 72.0 $\pm$ 12.3 & 66.7 $\pm$ 4.3 \\
\hline
\multirow{1}{*}{\parbox{0.2 cm}{C}}  & 0.72 $\pm$ 0.09 & 0.39 $\pm$ 0.07 & 0.24 $\pm$ 0.07 & 0.35 $\pm$ 0.07 & 0.18 $\pm$ 0.06 & 61.1 $\pm$ 6.8 & 72.0 $\pm$ 9.4 & 66.6 $\pm$ 4.2 \\
\hline
\rowcolor{green!15}
\multirow{1}{*}{\parbox{0.2cm}{D}}  & 0.76 $\pm$ 0.08 & 0.39 $\pm$ 0.01 & 0.24 $\pm$ 0.06 & 0.37 $\pm$ 0.07 & 0.20 $\pm$ 0.05 & 61.1 $\pm$ 1.5 & 75.8 $\pm$ 7.8 & 68.5 $\pm$ 4.2 \\ 
\hline
\multicolumn{9}{c}{\textbf{\textcolor{orange}{Max Pooling Strategy}}} \\ 
\hline
\rowcolor{green!5}
\multirow{1}{*}{\parbox{0.9cm}{$A_{max}$}} & 0.79 $\pm$ 0.04 & 0.45 $\pm$ 0.05 & 0.23 $\pm$ 0.07 & 0.35 $\pm$ 0.08 & 0.17 $\pm$ 0.06 & 54.7 $\pm$ 5.5 & 79.2 $\pm$ 3.8 & 67.0 $\pm$ 3.5 \\
\hline
\multirow{1}{*}{\parbox{0.9cm}{$B_{max}$}} & 0.79 $\pm$ 0.11 & 0.46 $\pm$ 0.06 & 0.22 $\pm$ 0.06 & 0.34 $\pm$ 0.07 & 0.16 $\pm$ 0.05 & 53.6 $\pm$ 6.0 & 79.4 $\pm$ 11.1 & 66.5 $\pm$ 3.3 \\
\hline
\multirow{1}{*}{\parbox{0.9cm}{$C_{max}$}}  & 0.74 $\pm$ 0.05 & 0.40 $\pm$ 0.05 & 0.24 $\pm$ 0.06 & 0.36 $\pm$ 0.06 & 0.18 $\pm$ 0.04 & 60.2 $\pm$ 4.9 & 74.3 $\pm$ 5.2 & 67.3 $\pm$ 2.7 \\
\hline
\rowcolor{green!5}
\multirow{1}{*}{\parbox{0.9cm}{$D_{max}$}} & 0.68 $\pm$ 0.06 & 0.34 $\pm$ 0.07 & 0.26 $\pm$ 0.07 & 0.37 $\pm$ 0.07 & 0.20 $\pm$ 0.06 & 66.2 $\pm$ 7.3 & 68.0 $\pm$ 6.4 & 67.1 $\pm$ 2.6 \\
\hline
\multicolumn{9}{c}{\textbf{\textcolor{orange}{Kernel Size change}}} \\ 
\hline
\rowcolor{green!25}
\multirow{1}{*}{\parbox{0.2cm}{E}} & 0.69 $\pm$ 0.05 & 0.34 $\pm$ 0.07 & 0.26 $\pm$ 0.06 & 0.37 $\pm$ 0.07 & 0.21 $\pm$ 0.06 & 66.4 $\pm$ 6.5 & 68.8 $\pm$ 4.8 & 67.6 $\pm$ 3.3 \\
\hline
\multirow{1}{*}{\parbox{0.2cm}{F}} & 0.69 $\pm$ 0.14 & 0.35 $\pm$ 0.07 & 0.25 $\pm$ 0.07 & 0.36 $\pm$ 0.06 & 0.19 $\pm$ 0.05 & 65.1 $\pm$ 7.0 & 69.6 $\pm$ 13.4 & 67.3 $\pm$ 4.4 \\
\hline
\multicolumn{9}{c}{\textbf{\textcolor{orange}{Deeper 8-layer model}}} \\
\hline
\multirow{1}{*}{\parbox{0.2cm}{G}} & 0.76 $\pm$ 0.07 & 0.37 $\pm$ 0.07 & 0.26 $\pm$ 0.07 & 0.38 $\pm$ 0.08 & 0.22 $\pm$ 0.07 & 63.4 $\pm$ 7.5 & 75.6 $\pm$ 7.5 & 69.5 $\pm$ 4.4 \\
\hline
\end{tabular}}
\end{table}

\begin{table}[!ht]
\centering
\caption{Independent test-set performance of the selected PPG CNN (Model E).}
\label{PPG_train_test}
\scalebox{0.78}{
\begin{tabular}{|>{\centering}p{2.0 cm}|>{\centering}p{2.0 cm}|>{\centering}p{2.0cm} |>{\centering}p{2.0cm}|>{\centering}p{2.0cm}|>{\centering}p{2.0cm}|>{\centering}p{2.1cm}|>{\centering}p{2.1cm}|P{2.1 cm}|}
\hline
\multirow{3}{*}{Model} & \multicolumn{8}{c|}{Performance Measure } \\
\cline{2-9}
&  TPR & FPR & Precision & F1 Score & Cohen Kappa & \textbf{No COBE class} accuracy (\%) & \textbf{COBE class} accuracy (\%) & Accuracy (\%) \\
\hline   
\multirow{1}{*}{\parbox{0.2cm}{E}} & 0.61 & 0.30 & 0.36 & 0.46 & 0.24 & 69.6 & 60.7 & 65.1 \\
\hline
\end{tabular}}
\end{table}

\subsection{Multi-signal training: late fusion}
Table~\ref{multi_cnnLF} summarises five-fold cross-validation results for late-fusion multi-signal CNNs (selected single-signal subnetworks concatenated and passed to a shared classifier). Table~\ref{multi_cnn_testLF} reports independent test-set performance for these late-fusion models.

\begin{table}[h!]
\centering
\caption{Five-fold cross-validation performance (mean $\pm$ SD) for multi-signal CNNs using late fusion.}\label{multi_cnnLF}
\scalebox{0.72}{
\begin{tabular}{|>{\centering}p{2.9 cm}|>{\centering}p{2.0 cm}|>{\centering}p{2.0cm} |>{\centering}p{2.0cm}|>{\centering}p{2.0cm}|>{\centering}p{2.0cm}|>{\centering}p{2.3cm}|>{\centering}p{2.3cm}|P{2.1 cm}|}
\hline
\multirow{3}{*}{Signals} & \multicolumn{8}{c|}{Performance Measure }\\
 \cline{2-9}
 & TPR & FPR & Precision & F1 Score & Cohen Kappa & \textbf{No COBE class} accuracy (\%) & \textbf{COBE class} accuracy (\%) & Accuracy (\%) \\
\hline
\multirow{1}{*}{\parbox{1.6cm}{IP + ECG}} & 0.90 $\pm$ 0.04 & 0.18 $\pm$ 0.04 & 0.45 $\pm$ 0.08 & 0.60 $\pm$ 0.06 & 0.51 $\pm$ 0.07 & 81.8 $\pm$ 4.0 & 90.3 $\pm$ 4.2 & 86.0 $\pm$ 2.5 \\
\hline
\hline
\multirow{1}{*}{\parbox{1.4cm}{IP + PPG}} & 0.90 $\pm$ 0.07 & 0.19 $\pm$ 0.02 & 0.44 $\pm$ 0.06 & 0.59 $\pm$ 0.06 & 0.49 $\pm$ 0.05 & 81.0 $\pm$ 2.4 & 89.8 $\pm$ 7.3 & 85.4 $\pm$ 3.7 \\
\hline
\hline
\multirow{1}{*}{\parbox{1.9cm}{ECG + PPG}} & 0.72 $\pm$ 0.08 & 0.32 $\pm$ 0.08 & 0.28 $\pm$ 0.07 & 0.40 $\pm$ 0.08 & 0.24 $\pm$ 0.07 & 67.9 $\pm$ 8.0 & 71.8 $\pm$ 7.5 & 69.9 $\pm$ 3.3 \\
\hline
\hline   
\multirow{1}{*}{\parbox{2.5cm}{IP + ECG + PPG}} & 0.90 $\pm$ 0.06 & 0.19 $\pm$ 0.02 & 0.44 $\pm$ 0.08 & 0.59 $\pm$ 0.07 & 0.49 $\pm$ 0.07 & 81.3 $\pm$ 2.6 & 89.5 $\pm$ 6.1 & 85.4 $\pm$ 3.5 \\
\hline
\end{tabular}}
\end{table} 

\begin{table}[!ht]
\centering
\caption{Independent test-set performance of late-fusion multi-signal CNN models (held-out infants).}
\label{multi_cnn_testLF}
\scalebox{0.72}{
\begin{tabular}{|>{\centering}p{2.9 cm}|>{\centering}p{2.0 cm}|>{\centering}p{2.0cm} |>{\centering}p{2.0cm}|>{\centering}p{2.0cm}|>{\centering}p{2.0cm}|>{\centering}p{2.3cm}|>{\centering}p{2.3cm}|P{2.1 cm}|}
\hline
\multirow{3}{*}{Signals} & \multicolumn{8}{c|}{Performance Measure} \\
 \cline{2-9}
 & TPR & FPR & Precision & F1 Score & Cohen Kappa & \textbf{No COBE class} accuracy (\%) & \textbf{COBE class} accuracy (\%) & Accuracy (\%) \\
 \hline
\multirow{1}{*}{\parbox{1.6cm}{IP + ECG}} & 0.89 & 0.17 & 0.60 & 0.72 & 0.62 & 83.2 & 89.5 & 86.3 \\
\hline
\hline
\multirow{1}{*}{\parbox{1.5cm}{IP + PPG}} & 0.88 & 0.15 & 0.63 & 0.73 & 0.64 & 85.2 & 87.9 & 86.6 \\
\hline
\hline
\multirow{1}{*}{\parbox{1.9cm}{ECG + PPG}} & 0.64 & 0.26 & 0.41 & 0.50 & 0.32 & 73.7 & 64.2 & 69.0 \\
\hline
\hline   
\multirow{1}{*}{\parbox{2.6cm}{IP + ECG + PPG}} & 0.88 & 0.18 & 0.58 & 0.70 & 0.59 & 81.7 & 88.5 & 85.1 \\
\hline
\end{tabular}}
\end{table} 

\section{Supplemetary results: COBE detection using ResNets}
\label{appendix:supp_resnet}

\subsection{Training using individual signals}
To evaluate the capacity of different ResNet architectures for one-dimensional signal classification, we trained and validated models on individual input signals using three variants: ResNet-18, ResNet-34, and ResNet-50. While ResNet-50 uses bottleneck blocks to reduce computational cost and preserve representational power, it did not consistently outperform the shallower variants. 

\begin{table}[!ht]
\centering
\caption{Cross-validation performance of ResNet models trained on individual signals. Results are reported as mean~$\pm$ standard deviation across five folds. For each signal type, the best-performing ResNet depth (ResNet-18, ResNet-34, or ResNet-50) is highlighted in bold.}
\label{resnet_cv}
\scalebox{0.77}{
\begin{tabular}{|>{\centering}p{1.9cm}|>{\centering}p{2.0 cm}|>{\centering}p{2.0cm}|>{\centering}p{2.0cm} |>{\centering}p{2.0cm}|>{\centering}p{2.0cm}|>{\centering}p{2.0cm}|>{\centering}p{1.9cm}|>{\centering}p{1.9cm}|P{1.8 cm}|}
\hline
\multirow{4}{*}{Signal} & \multirow{4}{*}{\parbox{1.8cm}{ResNet architecture}} & \multicolumn{8}{c|}{Performance Measure} \\
\cline{3-10}
& & TPR & FPR & Precision & F1 Score & Cohen Kappa & \textbf{No COBE class} accuracy (\%) & \textbf{COBE class} accuracy (\%) & Accuracy (\%) \\
\hline   
\multirow{3}{*}{\parbox{0.2cm}{IP}} & 18 & 0.87 $\pm$ 0.04 & 0.20 $\pm$ 0.04 & 0.43 $\pm$ 0.07 & 0.57 $\pm$ 0.07 & 0.46 $\pm$ 0.06 & 80.3 $\pm$ 4.1 & 86.4 $\pm$ 3.4 & 83.4 $\pm$ 1.7 \\ 
\cline{2-10}
& \textbf{34*} & \textbf{0.88 $\pm$ 0.04} & \textbf{0.19 $\pm$ 0.04} & \textbf{0.45 $\pm$ 0.05} & \textbf{0.59 $\pm$ 0.05} & \textbf{0.49 $\pm$ 0.04} & \textbf{81.6 $\pm$ 3.9} & \textbf{87.8 $\pm$ 4.0} & \textbf{84.7 $\pm$ 1.5} \\ 
\cline{2-10}
& 50 & 0.89 $\pm$ 0.07 & 0.19 $\pm$ 0.05 & 0.44 $\pm$ 0.09 & 0.58 $\pm$ 0.07 & 0.49 $\pm$ 0.08 & 80.9 $\pm$ 4.9 & 89.3 $\pm$ 7.2 & 85.1 $\pm$ 2.8 \\ 
\hline
\hline   
\multirow{3}{*}{\parbox{1.1cm}{$ECG_{edr}$}} & 18 & 0.70 $\pm$ 0.11 & 0.33 $\pm$ 0.01 & 0.26 $\pm$ 0.08 & 0.38 $\pm$ 0.09 & 0.22 $\pm$ 0.08 & 67.4 $\pm$ 1.4 & 69.8 $\pm$ 11.4 & 68.6 $\pm$ 5.8 \\ 
\cline{2-10}
& 34 & 0.73 $\pm$ 0.11 & 0.32 $\pm$ 0.03 & 0.28 $\pm$ 0.08 & 0.40 $\pm$ 0.09 & 0.25 $\pm$ 0.09 & 68.6 $\pm$ 3.3 & 73.2 $\pm$ 11.4 & 70.9 $\pm$ 6.3 \\ 
\cline{2-10}
& \textbf{50*} & \textbf{0.72 $\pm$ 0.11} & \textbf{0.28 $\pm$ 0.02} & \textbf{0.30 $\pm$ 0.08} & 0\textbf{.42 $\pm$ 0.09} & \textbf{0.28 $\pm$ 0.08} & \textbf{72.1 $\pm$ 2.3} & \textbf{71.7 $\pm$ 11.2} & \textbf{71.9 $\pm$ 5.5} \\ 
\hline   
\multirow{3}{*}{\parbox{1.1cm}{$ECG_{raw}$}} & \textbf{18*} & \textbf{0.60 $\pm$ 0.36} & \textbf{0.44 $\pm$ 0.29} & \textbf{0.16 $\pm$ 0.11} & \textbf{0.25 $\pm$ 0.16} &\textbf{ 0.08 $\pm$ 0.09} & \textbf{56.4 $\pm$ 28.7} & \textbf{59.6 $\pm$ 35.8} & \textbf{58.0 $\pm$ 7.0} \\ 
\cline{2-10}
& 34 & 0.35 $\pm$ 0.31 & 0.23 $\pm$ 0.25 & 0.20 $\pm$ 0.13 & 0.21 $\pm$ 0.14 & 0.09 $\pm$ 0.08 & 76.8 $\pm$ 25.4 & 34.8 $\pm$ 31.2 & 55.8 $\pm$ 4.2 \\ 
\cline{2-10}
& 50 & 0.41 $\pm$ 0.41 & 0.39 $\pm$ 0.40 & 0.12 $\pm$ 0.10 & 0.17 $\pm$ 0.16 & 0.01 $\pm$ 0.05 & 61.4 $\pm$ 39.6 & 41.0 $\pm$ 40.5 & 51.2 $\pm$ 3.5 \\ 
\hline
\hline   
\multirow{3}{*}{\parbox{1.1cm}{$PPG_{env}$}} & 18  & 0.76 $\pm$ 0.10 & 0.41 $\pm$ 0.07 & 0.24 $\pm$ 0.07 & 0.36 $\pm$ 0.08 & 0.18 $\pm$ 0.07 & 58.8 $\pm$ 6.8 & 75.9 $\pm$ 9.6 & 67.3 $\pm$ 4.3 \\ 
\cline{2-10}
& \textbf{34*}& \textbf{0.66 $\pm$ 0.09} & \textbf{0.33 $\pm$ 0.07} & \textbf{0.25 $\pm$ 0.04} & \textbf{0.36 $\pm$ 0.06} & \textbf{0.19 $\pm$ 0.06} & \textbf{66.4 $\pm$ 6.9} & \textbf{66.1 $\pm$ 8.8} & \textbf{66.3 $\pm$ 4.7} \\ 
\cline{2-10}
& 50 & 0.72 $\pm$ 0.14 & 0.37 $\pm$ 0.06 & 0.24 $\pm$ 0.06 & 0.36 $\pm$ 0.08 & 0.19 $\pm$ 0.06 & 62.8 $\pm$ 6.1 & 72.0 $\pm$ 14.2 & 67.4 $\pm$ 5.5 \\ 
\hline   
\multirow{3}{*}{\parbox{1.1cm}{$PPG_{raw}$}} &  18 & 0.70 $\pm$ 0.13 & 0.39 $\pm$ 0.10 & 0.24 $\pm$ 0.10 & 0.35 $\pm$ 0.11 & 0.17 $\pm$ 0.10 & 60.6 $\pm$ 10.7 & 70.4 $\pm$ 12.7 & 65.5 $\pm$ 4.9 \\ 
\cline{2-10}
& \textbf{34*} & \textbf{0.61 $\pm$ 0.09} & \textbf{0.30 $\pm$ 0.07 }& \textbf{0.26 $\pm$ 0.09} & \textbf{0.36 $\pm$ 0.08} & \textbf{0.21 $\pm$ 0.08} & \textbf{70.5 $\pm$ 7.5} & \textbf{61.3 $\pm$ 8.9} & \textbf{66.0 $\pm$ 3.3} \\
\cline{2-10}
& 50 & 0.67 $\pm$ 0.08 & 0.37 $\pm$ 0.10 & 0.24 $\pm$ 0.09 & 0.35 $\pm$ 0.10 & 0.17 $\pm$ 0.09 & 62.9 $\pm$ 9.6 & 66.7 $\pm$ 8.0 & 64.8 $\pm$ 4.1 \\
\hline   
\end{tabular}}
\end{table}

\begin{table}[!ht]
\centering
\caption[Test set performance of the selected ResNet models.]{Test set performance of the best-perfoming ResNet models identified during five-fold cross-validation (bolded in Table~\ref{resnet_cv}). Each model was trained on 83.3\% of the dataset (20 infants) and evaluated on the remaining 16.7\% (4 infants). Results reflect the final generalisation ability on held-out data.}
\label{resnet_test}
\scalebox{0.77}{
\begin{tabular}{|>{\centering}p{2.0cm}|>{\centering}p{1.8cm}|>{\centering}p{1.8cm} |>{\centering}p{1.8cm}|>{\centering}p{1.8cm}|>{\centering}p{1.8cm}|>{\centering}p{2.0cm}|>{\centering}p{2.0cm}|P{2.0 cm}|}
\hline
\multirow{3}{*}{Signal} & \multicolumn{8}{c|}{Performance Measure } \\
 \cline{2-9}
 & TPR & FPR & Precision & F1 Score & Cohen Kappa & \textbf{No COBE class} accuracy (\%) & \textbf{COBE class} accuracy (\%) & Accuracy (\%) \\
\hline
\multirow{1}{*}{\parbox{0.2cm}{IP}} & \textbf{0.93} & \textbf{0.17} & \textbf{0.61} & \textbf{0.73} & \textbf{0.63} & \textbf{82.8} & \textbf{92.7} & \textbf{87.7} \\  
\hline
\hline
\multirow{1}{*}{\parbox{1.1cm}{$ECG_{edr}$}} & \textbf{0.70} & \textbf{0.34} & \textbf{0.37} & \textbf{0.49} & \textbf{0.28} & \textbf{66.1} & \textbf{70.3} & \textbf{68.2} \\ 
\hline
\multirow{1}{*}{\parbox{1.1cm}{$ECG_{raw}$}} & 0.50 & 0.20 & 0.42 & 0.46 & 0.28 & 80.2 & 49.8 & 65.0 \\ 
\hline
\hline
\multirow{1}{*}{\parbox{1.1cm}{$PPG_{env}$}} & 0.63 & 0.31 & 0.37 & 0.47 & 0.26 & 69.5 & 62.6 & 66.1 \\ 
\hline
\multirow{1}{*}{\parbox{1.1cm}{$PPG_{raw}$}} & \textbf{0.58} & \textbf{0.26} & \textbf{0.39} & \textbf{0.46} & \textbf{0.27} & \textbf{74.2} & \textbf{57.5} & \textbf{65.8} \\ 
\hline
 \end{tabular}}
\end{table}

\subsection{Multi-modal training}
To assess whether performance could be improved by incorporating complementary signal information, we extended the training setup to include multiple inputs. Multi-modal learning is particularly relevant in COBE detection, where different signals capture distinct physiological responses to the cessation of breathing. We evaluated two common fusion strategies: late fusion and early fusion.

\subsubsection{Late Fusion}
In the late fusion configuration, each input signal is processed independently by its own ResNet encoder. The extracted feature representations are then concatenated and passed to a multi-layer perceptron for classification. The results of the late fusion experiments, shown in Tables~\ref{multi_resLF} and~\ref{multi_res_testLF}, compare performance across different combinations of input signals and ResNet depths. 

\begin{table}[!ht]
\centering
\caption{Cross-validation performance of ResNet  models using late fusion across different signal combinations. Results are reported as mean~$\pm$~standard deviation across five folds.}
\label{multi_resLF}
\scalebox{0.80}{
\begin{tabular}{|>{\centering}p{2.6 cm}|>{\centering}p{2.0 cm}|>{\centering}p{2.0cm} |>{\centering}p{2.0cm}|>{\centering}p{2.0cm}|>{\centering}p{2.0cm}|>{\centering}p{2.3cm}|>{\centering}p{2.3cm}|P{2.1 cm}|}
\hline
\multirow{3}{*}{Signals} & \multicolumn{8}{c|}{Performance Measure } \\
\cline{2-9}
 & TPR & FPR & Precision & F1 Score & Cohen Kappa & \textbf{No COBE class} accuracy (\%) & \textbf{COBE class} accuracy (\%) & Accuracy (\%) \\
\hline   
 \multirow{1}{*}{\parbox{1.5cm}{IP + ECG}}  & 0.89 $\pm$ 0.07 & 0.18 $\pm$ 0.02 & 0.44 $\pm$ 0.07 & 0.58 $\pm$ 0.05 & 0.49 $\pm$ 0.04 & 81.4 $\pm$ 2.0 & 89.3 $\pm$ 7.2 & 85.3 $\pm$ 3.3 \\
\hline
\hline
\multirow{1}{*}{\parbox{1.5cm}{IP + PPG}} & 0.88 $\pm$ 0.06 & 0.18 $\pm$ 0.04 & 0.45 $\pm$ 0.06 & 0.59 $\pm$ 0.05 & 0.50 $\pm$ 0.04 & 82.0 $\pm$ 4.0 & 88.3 $\pm$ 6.0 & 85.2 $\pm$ 2.6 \\ 
\hline
\hline
\multirow{1}{*}{\parbox{1.9cm}{ECG + PPG}} & 0.62 $\pm$ 0.09 & 0.27 $\pm$ 0.05 & 0.28 $\pm$ 0.07 & 0.39 $\pm$ 0.07 & 0.23 $\pm$ 0.07 & 73.4 $\pm$ 5.1 & 61.9 $\pm$ 8.8 & 67.7 $\pm$ 5.1 \\ 
\hline
\hline
\multirow{1}{*}{\parbox{2.5cm}{IP + ECG + PPG}}  & 0.89 $\pm$ 0.03 & 0.21 $\pm$ 0.06 & 0.43 $\pm$ 0.07 & 0.57 $\pm$ 0.06 & 0.47 $\pm$ 0.07 & 79.3 $\pm$ 6.4 & 89.5 $\pm$ 2.5 & 84.4 $\pm$ 2.6 \\
\hline   
\end{tabular}}
\end{table}

\begin{table}[!ht]
\centering
\caption[Test set performance of late fusion ResNet models.]{Test set performance of the best late fusion ResNet  models trained on different signal combinations. Each configuration uses separate ResNet encoders per signal, followed by feature concatenation and classification. These results reflect model generalisation to unseen data.}
\label{multi_res_testLF}
\scalebox{0.80}{
\begin{tabular}{|>{\centering}p{2.6 cm}|>{\centering}p{2.0 cm}|>{\centering}p{2.0cm} |>{\centering}p{2.0cm}|>{\centering}p{2.0cm}|>{\centering}p{2.0cm}|>{\centering}p{2.3cm}|>{\centering}p{2.3cm}|P{2.1 cm}|}
\hline
\multirow{3}{*}{Signals} & \multicolumn{8}{c|}{Performance Measure } \\
 \cline{2-9}
 &  TPR & FPR & Precision & F1 Score & Cohen Kappa & \textbf{No COBE class} accuracy (\%) & \textbf{COBE class} accuracy (\%) & Balanced Accuracy (\%) \\
\hline   
\multirow{1}{*}{\parbox{1.5cm}{IP + ECG}} & 0.92 & 0.20 & 0.56 & 0.70 & 0.58 & 79.6 & 92.0 & 85.8 \\ 
\hline
\hline   
\multirow{1}{*}{\parbox{1.5cm}{IP + PPG}}  & 0.91 & 0.17 & 0.60 & 0.73 & 0.63 & 82.8 & 91.4 & 87.1 \\
\hline
\hline   
\multirow{1}{*}{\parbox{1.9cm}{ECG + PPG}} & 0.50 & 0.18 & 0.44 & 0.47 & 0.30 & 81.9 & 49.8 & 65.9 \\
\hline
\hline   
\multirow{1}{*}{\parbox{2.5cm}{IP + ECG + PPG}} & 0.91 & 0.18 & 0.59 & 0.71 & 0.61 & 81.6 & 91.1 & 86.3 \\
\hline
\end{tabular}}
\end{table}

\newpage
\subsubsection{Early Fusion}
In the early fusion setup, all input signals are stacked channel-wise and passed through a shared ResNet encoder. This forces the network to learn joint representations from the beginning, capturing low-level interactions across modalities during the early stages of feature extraction.Tables~\ref{multi_resEF} and~\ref{multi_res_testEF} present the results for early fusion models trained on the full signal combination (IP, ECG, and PPG). 

\begin{table}[!ht]
\centering
\caption{Cross-validation performance of ResNet  models using early fusion on combined input signals. Results are reported as mean~$\pm$~standard deviation across five folds.}
\label{multi_resEF}
\scalebox{0.83}{
\begin{tabular}{|>{\centering}p{2.6 cm}|>{\centering}p{2.2 cm}|>{\centering}p{1.8 cm}|>{\centering}p{1.8 cm} |>{\centering}p{1.8 cm}|>{\centering}p{1.8 cm}|>{\centering}p{1.8 cm}|>{\centering}p{1.8cm}|>{\centering}p{1.8cm}|P{1.8 cm}|}
\hline
\multirow{5}{*}{Signals} & \multirow{5}{*}{\parbox{1.8cm}{ResNet architecture}} & \multicolumn{8}{c|}{Performance Measure } \\
\cline{3-10}
& & TPR & FPR & Precision & F1 Score & Cohen Kappa & \textbf{No COBE class} accuracy (\%) & \textbf{COBE class} accuracy (\%) & Accuracy (\%) \\
\hline
\multirow{3}{*}{\parbox{2.5cm}{IP + ECG + PPG}} & 18 & 0.90 $\pm$ 0.03 & 0.19 $\pm$ 0.03 & 0.45 $\pm$ 0.06 & 0.60 $\pm$ 0.06 & 0.50 $\pm$ 0.06 & 81.4 $\pm$ 3.1 & 89.9 $\pm$ 2.9 & 85.7 $\pm$ 1.9 \\
\cline{2-10}
& 34 & 0.89 $\pm$ 0.05 & 0.19 $\pm$ 0.04 & 0.44 $\pm$ 0.11 & 0.58 $\pm$ 0.09 & 0.49 $\pm$ 0.09 & 81.1 $\pm$ 4.5 & 89.2 $\pm$ 5.0 & 85.1 $\pm$ 2.7 \\
\cline{2-10}
& 50 & 0.87 $\pm$ 0.03 & 0.18 $\pm$ 0.04 & 0.45 $\pm$ 0.08 & 0.59 $\pm$ 0.07 & 0.49 $\pm$ 0.07 & 81.8 $\pm$ 3.5 & 87.2 $\pm$ 2.7 & 84.5 $\pm$ 2.4 \\
\hline
\end{tabular}}
\end{table}

\begin{table}[!ht]
\centering
\caption[Test set performance of early fusion ResNet models.]{Test set performance of early fusion ResNet models trained on a combination of IP, ECG, and PPG signals. Results reflect the model's ability to generalise to unseen data across varying ResNet depths.}
\label{multi_res_testEF}
\scalebox{0.83}{
\begin{tabular}{|>{\centering}p{2.4 cm}|>{\centering}p{2.2 cm}|>{\centering}p{1.8 cm}|>{\centering}p{1.8 cm} |>{\centering}p{1.8 cm}|>{\centering}p{1.8 cm}|>{\centering}p{1.8 cm}|>{\centering}p{1.8cm}|>{\centering}p{1.8cm}|P{1.8 cm}|}
\hline
\multirow{5}{*}{Signals} &\multirow{5}{*}{\parbox{1.8cm}{ResNet architecture}} & \multicolumn{8}{c|}{Performance Measure } \\
  \cline{3-10}
 & & TPR & FPR & Precision & F1 Score & Cohen Kappa & \textbf{No COBE class} accuracy (\%) & \textbf{COBE class} accuracy (\%) & Accuracy (\%) \\
\hline   
\multirow{3}{*}{\parbox{2.5cm}{IP + ECG + PPG}} & 18 & 0.85 & 0.12 & 0.67 & 0.75 & 0.66 & 87.7 & 85.3 & 86.5 \\
  \cline{2-10}
 & 34 & 0.91 & 0.18 & 0.59 & 0.72 & 0.62 & 82.0 & 91.4 & 86.7 \\
 \cline{2-10}
 & 50 & 0.90 & 0.19 & 0.58 & 0.71 & 0.60 & 81.4 & 89.8 & 85.6 \\ 
\hline
\end{tabular}}
\end{table}

\newpage
\section{Supplementary results: COBE detection using a ConvNeXt}
\label{appendix:supp_convnext}

\subsection{Individual signals}
Tables~\ref{conv_cv} and~\ref{conv_train_test} show the performances of the ConvNeXt model for individual input signals across different experimental settings. Specifically, we tested the ConvNeXt model using default settings, larger convolutional kernels, and reduced feature dimensionality. This comparison enabled us to investigate how these design choices affect classification performance across different signal types. The results presented in Table~\ref{conv_cv} show the model's behaviour on the cross-validation sets.

\begin{table}[!ht]
\centering
\caption[Cross-validation performance of ConvNeXt models (individual input signals)]{Cross-validation performance of ConvNeXt models for individual signals using different architectural configurations: default, larger kernels, and reduced feature dimensionality. Bolded entries indicate the best-performing setup per signal.}
\label{conv_cv}
\scalebox{0.80}{
\begin{tabular}{|>{\centering}p{1.9 cm}|>{\centering}p{2.4 cm}|>{\centering}p{1.9 cm}|>{\centering}p{1.9cm} |>{\centering}p{1.9cm}|>{\centering}p{1.9cm}|>{\centering}p{1.9cm}|>{\centering}p{1.9cm}|>{\centering}p{1.9cm}|P{1.8 cm}|}
\hline
\multirow{4}{*}{Signal} & \multirow{4}{*}{\parbox{1.8cm}{ConvNeXt architecture}} & \multicolumn{8}{c|}{Performance Measure } \\
 \cline{3-10}
& & TPR & FPR & Precision & F1 Score & Cohen Kappa & \textbf{No COBE class} accuracy (\%) & \textbf{COBE class} accuracy (\%) & Accuracy (\%) \\
\hline   
  \multirow{5}{*}{\parbox{0.3cm}{IP}} & Default& 0.90 $\pm$ 0.02 & 0.18 $\pm$ 0.05 & 0.45 $\pm$ 0.07 & 0.60 $\pm$ 0.06 & 0.51 $\pm$ 0.07 & 81.8 $\pm$ 5.0 & 89.7 $\pm$ 1.8 & 85.7 $\pm$ 2.2 \\ 
  \cline{2-10}
& \textbf{Larger Kernels}  & \textbf{0.89 $\pm$ 0.02} & \textbf{0.16 $\pm$ 0.05} & \textbf{0.49 $\pm$ 0.09} & \textbf{0.63 $\pm$ 0.07} & \textbf{0.54 $\pm$ 0.08} & \textbf{84.2 $\pm$ 4.5} & \textbf{89.1 $\pm$ 1.9} & \textbf{86.7 $\pm$ 1.9 }\\ 
 \cline{2-10}
& Reduced Features & \multirow{2}{*}{0.89 $\pm$ 0.03} & \multirow{2}{*}{0.17 $\pm$ 0.04} & \multirow{2}{*}{0.48 $\pm$ 0.09} & \multirow{2}{*}{0.62 $\pm$ 0.07} & \multirow{2}{*}{0.53 $\pm$ 0.08} & \multirow{2}{*}{83.5 $\pm$ 4.1} & \multirow{2}{*}{88.8 $\pm$ 2.7} & \multirow{2}{*}{86.2 $\pm$ 2.1} \\
\hline
\hline   
\multirow{5}{*}{\parbox{1.2cm}{$ECG_{edr}$}} & \textbf{Default} & \textbf{0.82 $\pm$ 0.07} & \textbf{0.39 $\pm$ 0.06} & \textbf{0.26 $\pm$ 0.06} & \textbf{0.40 $\pm$ 0.07} & \textbf{0.23 $\pm$ 0.07} & \textbf{61.5 $\pm$ 5.6} & \textbf{82.3 $\pm$ 7.3} & \textbf{71.9 $\pm$ 5.5} \\ 
\cline{2-10}
& Larger Kernels & 0.81 $\pm$ 0.05 &0.40 $\pm$ 0.08 &0.26 $\pm$ 0.06 & 0.39 $\pm$ 0.07 & 0.21 $\pm$ 0.07 & 60.1 $\pm$ 8.2 & 81.1 $\pm$ 5.1 & 70.6 $\pm$ 5.4 \\
\cline{2-10}
& Reduced Features & \multirow{2}{*}{0.82 $\pm$ 0.09} & \multirow{2}{*}{0.38 $\pm$ 0.05} & \multirow{2}{*}{0.26 $\pm$ 0.06} &\multirow{2}{*}{0.40 $\pm$ 0.07} & \multirow{2}{*}{0.23 $\pm$ 0.07} & \multirow{2}{*}{62.1 $\pm$ 5.2} & \multirow{2}{*}{81.7 $\pm$ 9.0} & \multirow{2}{*}{71.9 $\pm$ 6.0} \\
 \hline   
\multirow{5}{*}{\parbox{1.2cm}{$ECG_{raw}$}} & Default & 0.77 $\pm$ 0.07 & 0.57 $\pm$ 0.12 & 0.19 $\pm$ 0.05 & 0.30 $\pm$ 0.06 & 0.09 $\pm$ 0.04 & 42.7 $\pm$ 11.7 & 77.0 $\pm$ 7.5 & 59.9 $\pm$ 4.3 \\
\cline{2-10}
 & Larger Kernels & 0.79 $\pm$ 0.07 & 0.62 $\pm$ 0.11 & 0.18 $\pm$ 0.06 &0.29 $\pm$ 0.07 &0.07 $\pm$ 0.04 & 37.7 $\pm$ 11.4 & 78.7 $\pm$ 6.6 & 58.2 $\pm$ 3.8 \\
\cline{2-10}
 & \textbf{Reduced Features} & \multirow{2}{*}{\textbf{0.77 $\pm$ 0.11}} & \multirow{2}{*}{\textbf{0.54 $\pm$ 0.20}} & \multirow{2}{*}{\textbf{0.20 $\pm$ 0.07}} & \multirow{2}{*}{\textbf{0.31 $\pm$ 0.08}} & \multirow{2}{*}{\textbf{0.11 $\pm$ 0.07 }}& \multirow{2}{*}{\textbf{46.0 $\pm$ 19.8}} & \multirow{2}{*}{\textbf{76.7 $\pm$ 11.0}} & \multirow{2}{*}{\textbf{61.4 $\pm$ 5.6}} \\
\hline
\hline   
\multirow{5}{*}{\parbox{1.2cm}{$PPG_{env}$}} & \textbf{Default} & \textbf{0.89 $\pm$ 0.06} & \textbf{0.50 $\pm$ 0.03} & \textbf{0.23 $\pm$ 0.06} & \textbf{0.36 $\pm$ 0.08} & \textbf{0.18 $\pm$ 0.05} & \textbf{50.3 $\pm$ 3.0} & \textbf{88.7 $\pm$ 6.3} & \textbf{69.5 $\pm$ 3.6} \\
\cline{2-10}
& Larger Kernels & 0.82 $\pm$ 0.05 & 0.48 $\pm$ 0.03 & 0.22 $\pm$ 0.06 & 0.35 $\pm$ 0.08 & 0.16 $\pm$ 0.05 & 51.8 $\pm$ 3.2 & 81.9 $\pm$ 4.9 & 66.9 $\pm$ 3.5 \\
\cline{2-10}
& Reduced Features & \multirow{2}{*}{0.84 $\pm$ 0.07} & \multirow{2}{*}{0.48 $\pm$ 0.04} & \multirow{2}{*}{0.23 $\pm$ 0.06} & \multirow{2}{*}{0.35 $\pm$ 0.08} & \multirow{2}{*}{0.17 $\pm$ 0.06} & \multirow{2}{*}{51.5 $\pm$ 3.7} & \multirow{2}{*}{84.3 $\pm$ 7.0} & \multirow{2}{*}{67.9 $\pm$ 4.5} \\
\hline   
\multirow{5}{*}{\parbox{1.2cm}{$PPG_{raw}$}} & Default & 0.70 $\pm$ 0.12 & 0.42 $\pm$ 0.07 & 0.22 $\pm$ 0.08 & 0.33 $\pm$ 0.09 & 0.15 $\pm$ 0.08 & 57.9 $\pm$ 7.0 & 70.3 $\pm$ 11.6 & 64.1 $\pm$ 6.1 \\ 
\cline{2-10}
& \textbf{{Larger Kernels}} & \textbf{0.71 $\pm$ 0.07} & \textbf{0.39 $\pm$ 0.05} & \textbf{0.24 $\pm$ 0.07} & \textbf{0.35 $\pm$ 0.08} & \textbf{0.17 $\pm$ 0.07} & \textbf{60.9 $\pm$ 5.3} & \textbf{70.7 $\pm$ 6.9} & \textbf{65.8 $\pm$ 5.4}\\
\cline{2-10}
 & Reduced Features & \multirow{2}{*}{0.71 $\pm$ 0.12} & \multirow{2}{*}{0.44 $\pm$ 0.08} & \multirow{2}{*}{0.22 $\pm$ 0.08} & \multirow{2}{*}{0.33 $\pm$ 0.09} & \multirow{2}{*}{0.14 $\pm$ 0.07} & \multirow{2}{*}{56.6 $\pm$ 8.3} & \multirow{2}{*}{70.8 $\pm$ 11.8} & \multirow{2}{*}{63.7 $\pm$ 4.6} \\
\hline
 \end{tabular}}
\end{table}

\begin{table}[!ht]
\centering
\caption[Test performance for COBE detection using ConvNeXt models]{Test performance for COBE detection using ConvNeXt models trained on various physiological signals. Each row presents performance metrics for a specific signal-modality and model configuration. Results reflect each model's ability to generalise to unseen data.}
\label{conv_train_test}
\scalebox{0.80}{
\begin{tabular}{|>{\centering}p{2.0 cm}|>{\centering}p{1.8cm}|>{\centering}p{1.8cm} |>{\centering}p{1.8cm}|>{\centering}p{1.8cm}|>{\centering}p{1.8cm}|>{\centering}p{2.0cm}|>{\centering}p{2.0cm}|P{2.0 cm}|}
\hline
\multirow{3}{*}{Signal} & \multicolumn{8}{c|}{Performance Measure } \\
 \cline{2-9}
& TPR & FPR & Precision & F1 Score & Cohen Kappa & \textbf{No COBE class} accuracy (\%) & \textbf{COBE class} accuracy (\%) & Accuracy (\%) \\
\hline   
IP & \textbf{0.91} & \textbf{0.15} & \textbf{0.64} & \textbf{0.75} & \textbf{0.66} & \textbf{85.3} & \textbf{90.7 }& \textbf{88.0} \\
\hline
\hline   
$ECG_{edr}$ & \textbf{0.77} & \textbf{0.37} & \textbf{0.37} & \textbf{0.50} & \textbf{0.29} & \textbf{62.7} & \textbf{76.7} & \textbf{69.7} \\ 
\hline   
$ECG_{raw}$ & 0.58 & 0.34 & 0.33 & 0.42 & 0.19 & 66.2 & 57.8 & 62.0 \\
 \hline
\hline   
$PPG_{env}$ & \textbf{0.73} & \textbf{0.40} & \textbf{0.34} & \textbf{0.47} & \textbf{0.23} & \textbf{60.0} & \textbf{72.8} & \textbf{66.4} \\
\hline   
$PPG_{raw}$ & 0.67 & 0.44 & 0.31 & 0.42 & 0.17 & 56.4 & 67.4 & 61.9 \\
\hline
 \end{tabular}}
\end{table}

\newpage
\subsection{Multi-modal training}
To assess whether performance could be improved by incorporating complementary signal information, we extended the ConvNeXt training setup to support multiple input modalities. Multimodal learning is particularly relevant in COBE detection, where different signals capture distinct physiological responses to cessation of breathing. We evaluated two common fusion strategies: late fusion, which combines the outputs of independently trained single-signal models and early fusion, which combines multiple signals at the input level. 

\subsubsection{Late Fusion}
In the late fusion approach, features were extracted from each ConvNeXt branch before their classification layers, concatenated, and then passed through a separate multi-layer perceptron (MLP) for final classification. Tables~\ref{multi_convLF} and \ref{multi_conv_testLF} summarise the results from late fusion during training (with cross-validation) and on held-out test data, respectively.

\begin{table}[!ht]
\centering
\caption[Cross-validation results for late fusion ConvNeXt models with multiple signals.]{Cross-validation performance of late fusion ConvNeXt models trained on different combinations of physiological signals. Results are presented as mean and standard deviation across the five folds.}
\label{multi_convLF}
\scalebox{0.83}{
\begin{tabular}{|>{\centering}p{2.6 cm}|>{\centering}p{2.0 cm}|>{\centering}p{2.0cm} |>{\centering}p{2.0cm}|>{\centering}p{2.0cm}|>{\centering}p{2.0cm}|>{\centering}p{2.3cm}|>{\centering}p{2.3cm}|P{2.1 cm}|}
\hline
\multirow{3}{*}{Signals} & \multicolumn{8}{c|}{Performance Measure } \\
 \cline{2-9}
 & TPR & FPR & Precision & F1 Score & Cohen Kappa & \textbf{No COBE class} accuracy (\%) & \textbf{COBE class} accuracy (\%) & Accuracy (\%) \\

\hline   
 \multirow{1}{*}{\parbox{1.5cm}{IP + ECG}}  & 0.89 $\pm$ 0.01 & 0.19 $\pm$ 0.06 & 0.45 $\pm$ 0.08 & 0.59 $\pm$ 0.07 & 0.49 $\pm$ 0.09 & 80.7 $\pm$ 5.9 & 89.5 $\pm$ 1.2 & 85.1 $\pm$ 3.2 \\
\hline
\hline
 \multirow{1}{*}{\parbox{1.4cm}{IP + PPG}} & 0.89 $\pm$ 0.04 & 0.18 $\pm$ 0.05 & 0.46 $\pm$ 0.09 & 0.60 $\pm$ 0.07 & 0.51 $\pm$ 0.08 & 82.0 $\pm$ 5.0 & 89.3 $\pm$ 4.0 & 85.6 $\pm$ 3.3 \\
\hline
\hline
\multirow{1}{*}{\parbox{1.9cm}{ECG + PPG}} & 0.82 $\pm$ 0.05 & 0.38 $\pm$ 0.07 & 0.27 $\pm$ 0.05 & 0.40 $\pm$ 0.07 & 0.24 $\pm$ 0.06 & 62.3 $\pm$ 7.1 & 82.0 $\pm$ 4.5 & 72.2 $\pm$ 4.8 \\
\hline
\hline
\multirow{1}{*}{\parbox{2.5cm}{IP + ECG + PPG}}  & 0.90 $\pm$ 0.04 & 0.20 $\pm$ 0.05 & 0.44 $\pm$ 0.08 & 0.59 $\pm$ 0.07 & 0.49 $\pm$ 0.07 & 80.5 $\pm$ 5.2 & 90.3 $\pm$ 3.5 & 85.4 $\pm$ 3.0 \\
\hline   
 \end{tabular}}
\end{table}

\begin{table}[!ht]
\centering
\caption[Test performance of late fusion ConvNeXt models trained on various combinations of physiological signals.]{Test set performance of late fusion ConvNeXt models trained on various combinations of physiological signals. Metrics reflect each model's ability to generalise to unseen data.}
\label{multi_conv_testLF}
\scalebox{0.83}{
\begin{tabular}{|>{\centering}p{2.6 cm}|>{\centering}p{2.0 cm}|>{\centering}p{2.0cm} |>{\centering}p{2.0cm}|>{\centering}p{2.0cm}|>{\centering}p{2.0cm}|>{\centering}p{2.3cm}|>{\centering}p{2.3cm}|P{2.1 cm}|}
\hline
\multirow{3}{*}{Signals} & \multicolumn{8}{c|}{Performance Measure } \\
 \cline{2-9}
 & TPR & FPR & Precision & F1 Score & Cohen Kappa & \textbf{No COBE class} accuracy (\%) & \textbf{COBE class} accuracy (\%) & Accuracy (\%) \\
\hline   
  \multirow{1}{*}{\parbox{1.5cm}{IP + ECG}}  & 0.91 & 0.17 & 0.61 & 0.73 & 0.63 & 83.0 & 90.7 & 86.9 \\ 
 \hline
\hline   
  \multirow{1}{*}{\parbox{1.4cm}{IP + PPG}} & 0.94 & 0.16 & 0.62 & 0.75 & 0.66 & 83.9 & 93.6 & 88.7\\
\hline
\hline   
  \multirow{1}{*}{\parbox{1.9cm}{ECG + PPG}} & 0.80 & 0.37 & 0.38 & 0.51 & 0.31 & 62.9 & 79.6 & 71.2 \\
 \hline
\hline   
  \multirow{1}{*}{\parbox{2.5cm}{IP + ECG + PPG}} & 0.92 & 0.16 & 0.62 & 0.74 & 0.65 & 84.1 & 92.3 & 88.2 \\
\hline
 \end{tabular}}
\end{table}

\newpage
\subsection{Early Fusion}
The performance of early fusion ConvNeXt models was evaluated using different architectural variations on combined IP, ECG and PPG signals. Table \ref{multi_convEF} presents the cross-validation results, showing the average metrics and their standard deviations across five training folds. Table \ref{multi_conv_testEF} reports the corresponding test set performance, demonstrating the models' ability to generalise to unseen data. 

\begin{table}[!ht]
\centering
\caption[Cross-validation performance of early fusion ConvNeXt models on combined~IP, ECG and PPG signals]{Cross-validation performance of early fusion ConvNeXt models trained on combined~IP, ECG and PPG signals, showing mean and standard deviation for each metric across five folds.}
\label{multi_convEF}
\scalebox{0.83}{
\begin{tabular}{|>{\centering}p{2.6 cm}|>{\centering}p{2.2 cm}|>{\centering}p{1.8 cm}|>{\centering}p{1.8 cm} |>{\centering}p{1.8 cm}|>{\centering}p{1.8 cm}|>{\centering}p{1.8 cm}|>{\centering}p{1.8cm}|>{\centering}p{1.8cm}|P{1.8 cm}|}
\hline
\multirow{5}{*}{Signals} & \multirow{5}{*}{\parbox{1.8cm}{ConvNeXt architecture}} & \multicolumn{8}{c|}{Performance Measure } \\
 \cline{3-10}
 & & TPR & FPR & Precision & F1 Score & Cohen Kappa & \textbf{No COBE class} accuracy (\%) & \textbf{COBE class} accuracy (\%) & Accuracy (\%) \\

\hline   
   \multirow{4}{*}{\parbox{2.5cm}{IP + ECG + PPG}} & Default & 0.92 $\pm$ 0.02 & 0.20 $\pm$ 0.06 & 0.44 $\pm$ 0.10 & 0.59 $\pm$ 0.09 & 0.49 $\pm$ 0.10 & 79.6 $\pm$ 6.4 & 92.1 $\pm$ 1.9 & 85.9 $\pm$ 3.2 \\
   
  \cline{2-10}
 & Larger Kernels & 0.89 $\pm$ 0.03 & 0.19 $\pm$ 0.06 & 0.45 $\pm$ 0.09 & 0.59 $\pm$ 0.09 & 0.50 $\pm$ 0.10 & 81.4 $\pm$ 5.7 & 89.5 $\pm$ 3.0 & 85.4 $\pm$ 3.0 \\
 
 \cline{2-10}
 & Reduced Features & 0.91 $\pm$ 0.02 & 0.19 $\pm$ 0.05 & 0.44 $\pm$ 0.09 & 0.60 $\pm$ 0.08 & 0.50 $\pm$ 0.10 & 80.6 $\pm$ 5.3 & 90.8 $\pm$ 1.7 & 85.7 $\pm$ 3.3 \\
\hline
\end{tabular}}
\end{table}

\begin{table}[!ht]
\centering
\caption[Test performance of early fusion ConvNeXt models on combined IP, ECG and PPG signals.]{Test set performance of early fusion ConvNeXt models trained on combined IP, ECG and  PPG signals using different architectural variations.}

\label{multi_conv_testEF}
\scalebox{0.83}{
\begin{tabular}{|>{\centering}p{2.6 cm}|>{\centering}p{2.2 cm}|>{\centering}p{1.8 cm}|>{\centering}p{1.8cm} |>{\centering}p{1.8cm}|>{\centering}p{1.8cm}|>{\centering}p{1.8cm}|>{\centering}p{1.8cm}|>{\centering}p{1.8cm}|P{1.8cm}|}
\hline
\multirow{5}{*}{Signals} & \multirow{5}{*}{\parbox{1.8cm}{ConvNeXt architecture}} & \multicolumn{8}{c|}{Performance Measure } \\
 \cline{3-10} 
 & & TPR & FPR & Precision & F1 Score & Cohen Kappa & \textbf{No COBE class} accuracy (\%) & \textbf{COBE class} accuracy (\%) & Accuracy (\%) \\

\hline   
   \multirow{4}{*}{\parbox{2.5cm}{IP + ECG + PPG}} & Default & 0.88 & 0.14 & 0.64 & 0.74 & 0.65 & 85.6 & 87.9 & 86.7 \\
   
  \cline{2-10}
 & Larger Kernels & 0.93 & 0.19 & 0.59 & 0.72 & 0.61 & 81.3 & 92.7 & 87.0 \\
 
 \cline{2-10}
 & Reduced Features & 0.90 & 0.18 & 0.59 & 0.71 & 0.61 & 81.8 & 90.4 & 86.1 \\
\hline
\end{tabular}}
\end{table}

\newpage
\bibliography{references}

\end{document}